\documentclass[10pt,twocolumn,letterpaper]{article}

\usepackage[pagenumbers]{cvpr}              % To produce the CAMERA-READY version
\usepackage{graphicx}
\usepackage{amsmath}
\usepackage{amssymb}
\usepackage{booktabs}
\usepackage{times}
\usepackage{epsfig}
\usepackage{bm}
\usepackage{multirow}
\usepackage{cuted}
\usepackage{capt-of}
\usepackage{caption}
\usepackage{animate}
\usepackage{bbding}
\usepackage{footnote}
\usepackage{diagbox}
\usepackage{colortbl}
\usepackage{tabularx}
\usepackage{tikz}
\usepackage[toc,page]{appendix}
\usepackage[accsupp]{axessibility}

\definecolor{cvprblue}{rgb}{0.21,0.49,0.74}
\usepackage[pagebackref,breaklinks,colorlinks,citecolor=cvprblue]{hyperref}

\usepackage[capitalize]{cleveref}
\crefname{section}{Sec.}{Secs.}
\Crefname{section}{Section}{Sections}
\Crefname{table}{Table}{Tables}
\crefname{table}{Tab.}{Tabs.}

\graphicspath{{images/}}

\newcommand{\paravspace}{\vspace{-12pt}}
\newcommand{\tikzcircle}[2][red,fill=red]{\tikz[baseline=-0.7ex]\draw[#1,radius=#2] (0,0) circle ;}

\definecolor{gold}{RGB}{221, 196, 65}
\definecolor{silver}{RGB}{215, 215, 215}
\definecolor{bronze}{RGB}{126, 66, 5}

\graphicspath{{images/}}

\definecolor{yellow}{rgb}{1, 1, 0.7}
\definecolor{orange}{rgb}{1, 0.85, 0.7}
\definecolor{pink}{rgb}{1, 0.7, 0.7}

\def\paperID{*****} % *** Enter the Paper ID here
\def\confName{CVPR}
\def\confYear{2024}

\title{Portrait4D: Learning One-Shot 4D Head Avatar Synthesis using Synthetic Data}

\author{Yu Deng \quad Duomin Wang \quad Xiaohang Ren \quad Xingyu Chen \quad Baoyuan Wang\\
	Xiaobing.AI\\
 \small{\url{https://yudeng.github.io/Portrait4D/}}
}

\begin{document}
\maketitle

\begin{strip}
\vspace{-45pt}
	\centering
	\includegraphics[width=1.0\textwidth]{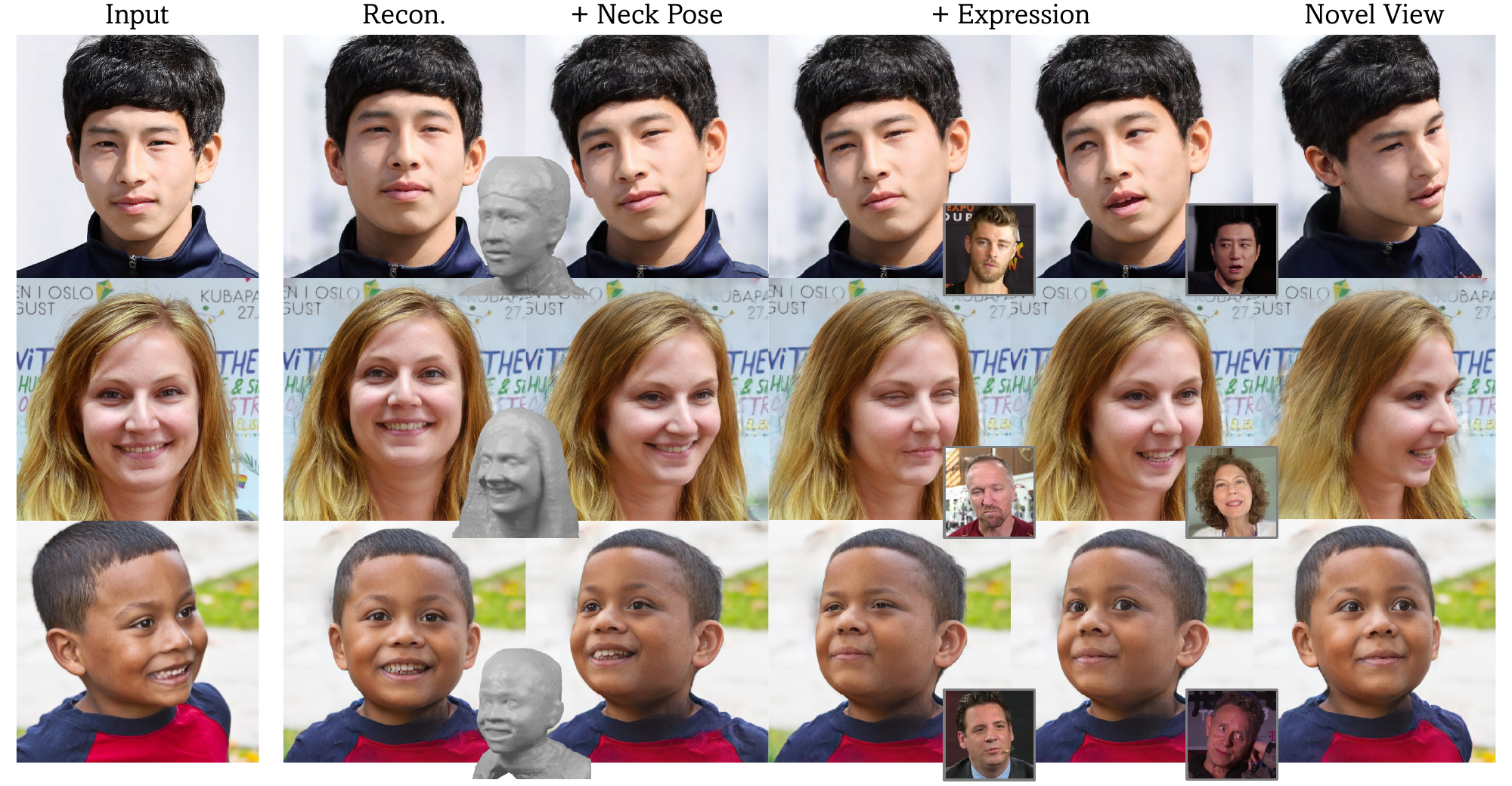}\\
	\captionsetup{type=figure,font=small}
	\vspace{-8pt}
	\caption{Our method generates a photorealistic 4D head avatar via a feed-forward process of a monocular image. It allows head reenactment given another driving image, with full motion control of face, mouth, eyes, and neck, as well as foreground-background separation.}
	\label{fig:teaser}
\end{strip}
\vspace{-4pt}

\begin{abstract}
Existing one-shot 4D head synthesis methods usually learn from monocular videos with the aid of 3DMM reconstruction, yet the latter is evenly challenging which restricts them from reasonable 4D head synthesis. We present a method to learn one-shot 4D head synthesis via large-scale synthetic data. The key is to first learn a part-wise 4D generative model from monocular images via adversarial learning, to synthesize multi-view images of diverse identities and full motions as training data; then leverage a transformer-based animatable triplane reconstructor to learn 4D head reconstruction using the synthetic data. A novel learning strategy is enforced to enhance the generalizability to real images by disentangling the learning process of 3D reconstruction and reenactment. Experiments demonstrate our superiority over the prior art.
\end{abstract}

%--------------------------------------------------------------------

\section{Introduction}
\label{sec:intro}
Animating a portrait image for photorealistic video synthesis (\ie, one-shot head avatar synthesis) is a crucial task in computer vision and graphics, which can benefit diverse applications like video conferencing, live streaming, and Virtual Reality. Compared to various approaches based on 2D generative models~\cite{siarohin2019first,burkov2020neural,wang2021one,yin2022styleheat,drobyshev2022megaportraits,zhang2023metaportrait}, animatable 3D head (\ie, 4D head) synthesis yields better 3D consistency during view changes, thus more favorable in scenarios that require large head pose variation and free-view rendering.

Nevertheless, reconstructing an animatable 3D head from a single image is highly challenging, and there lack of large-scale 3D data to directly learn a deep model to tackle it. Consequently, previous methods~\cite{xu2020deep,khakhulin2022realistic,ma2023otavatar,li2023one,yu2023nofa} resort to monocular videos and leverage differentiable renderers for training with image-level weak supervisions. Since this process is highly ill-posed, monocular 3DMM reconstructions~\cite{blanz1999morphable,paysan20093d,li2017learning,bulat2017far,deng2019accurate,feng2021learning} are often involved to extract pose, shape, or expression to facilitate the geometry learning. However, 3DMM reconstruction from a monocular image is evenly challenging, which hinders the existing methods to obtain reasonable geometries for large-pose reenactment. What's more, they often neglect eye gaze and neck pose control to simplify the 
problem. Besides, the monocular video setting also makes it difficult to learn foreground and background separation. 

Therefore, we wonder if it is possible to leverage synthetic data that can provide extensive multi-view images with diverse identities and expressions for training. With them, we can fully exploit the power of advanced neural networks~\cite{dosovitskiy2020image,xie2021segformer} to learn 4D head synthesis in a data-driven manner, avoiding errors from monocular 3DMM reconstruction. Background isolation is also feasible with separately synthesized foreground and background data. However, synthesizing such data with enough photorealism is extremely difficult. 
 A recent method~\cite{trevithick2023real} resorts to multi-view images generated by a 3D GAN~\cite{chan2021efficient} for learning static novel view synthesis, yet it does not deal with animations. And existing 3D GANs~\cite{bergman2022generative,wu2022anifacegan,wu2023aniportraitgan,xu2023omniavatar,sun2023next3d} are difficult to generate head images with full motion control and separated background that meet our requirements. What's more, it is unclear if a model learned on such synthetic data can generalize well to real image reenactment. 

In this paper, we present a framework for learning one-shot 4D head avatar synthesis by creating large-scale synthetic data as supervision. Learning on the synthetic data alone, our framework achieves high-fidelity 4D head reconstruction of real images via a feed-forward pass, and supports motion control of the face, mouth, eyes, and neck, given another driving image for reenactment. We also allow independent foreground animation with background separation. The core of our framework is two-fold: 
 \textbf{1)}, a 4D generative model (GenHead) capable of synthesizing photorealistic head images with free pose and full head motion control, trained on only monocular images. \textbf{2)}, a one-shot 4D head synthesis model (Portrait4D) together with its disentangled learning strategy for photorealistic 4D head reconstruction from a real image while trained on synthetic data of GenHead.

The key of GenHead is a combination of a tri-plane generator~\cite{chan2021efficient} for canonical head NeRF~\cite{mildenhall2020nerf} synthesis and a deformation field derived from FLAME meshes~\cite{li2017learning} for animation. Different from previous 3D head GANs~\cite{bergman2022generative,xu2023omniavatar}, we leverage part-wise deformation fields and canonical tri-planes to deal with complex motions around eyes and mouth, which help with full head motion control important to vivid reenactment. GenHead is trained on in-the-wild real images via image-level adversarial learning~\cite{goodfellow2014generative,chan2021efficient} and successfully ``turns" these monocular data to 4D synthetic ones to enable learning the 4D head reconstruction. 

Our 4D head synthesis model adopts a transformer-based encoder-decoder architecture which directly reconstructs a triplane-based head NeRF from a source image and animates it via deep motion features from a driving image. We utilize a pre-trained motion encoder~\cite{wang2023progressive} for expression extraction with identity information excluded, and inject it into the tri-plane reconstruction process via cross-attention with the appearance features to achieve motion control. 
This way, we get rid of the reliance on 3DMM estimation for pose and expression.
Our pipeline is trained end-to-end via self-reenacting the synthetic identities and comparing the corresponding results at arbitrary views with their ground truth. This differentiates us with previous methods leveraging a monocular frame as both the driving and the ground truth, and yields better 3D geometries.

While our synthetic data are of high photorealism, they still have domain gaps with the real ones which influence the model's generalizability. To tackle it, we further introduce a disentangled learning strategy that isolates the 3D reconstruction and the reenactment process of the model to alleviate overfitting. The core idea is to randomly switch off the motion-related cross-attention layers and let the remaining parts focus on static 3D reconstruction only. This way, the learned model achieves faithful 3D head reconstruction and animation on unseen real images.

We train our GenHead model on FFHQ dataset~\cite{karras2019style} at $512^2$ resolution and use it to generate synthetic 4D data to learn the subsequent 4D head synthesis model. Experiments show that our method achieves high-fidelity 4D head reconstruction with reasonable geometry and complete motion control (Fig.~\ref{fig:teaser}), outperforming previous approaches. We believe our method opens up a new way for scaling up photorealistic head avatar creation.

\begin{figure*}[t]
	\small
	\centering
	\includegraphics[width=0.95\textwidth]{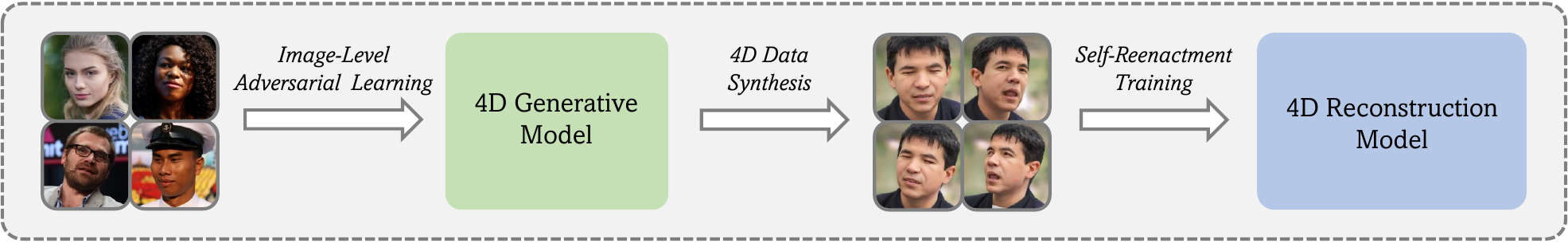}
    \vspace{-4pt}
	\caption{Overview of our method. We first learn a 4D generative head model from monocular images to synthesize large-scale 4D data. Then, we utilize the synthetic data to learn a one-shot 4D head reconstruction model in a data-driven manner.}
	\label{fig:overview}
 \vspace{-8pt}
\end{figure*}

%--------------------------------------------------------------------
\section{Related Work}
\label{sec:related}

\paragraph{One-shot head avatar synthesis.} One-shot head avatar synthesis aims to animate a source portrait via motions from a target image. 
In recent years, plenty of works have taken advantage of 2D generative models for talking head synthesis at high fidelity. For example, \cite{wiles2018x2face,siarohin2019first, wang2021one, ren2021pirenderer,hong2022depth,drobyshev2022megaportraits,zhang2023metaportrait,gao2023high} learn latent feature deformations and send the deformed features to a 2D generator for face reenactment. 
\cite{yin2022styleheat,guan2023stylesync,bounareli2023hyperreenact} map images to the latent space of a pre-trained StyleGAN2~\cite{karras2020analyzing} to utilize the power of the unconditional GAN. While these methods can produce photorealistic face images, they struggle to preserve the 3D structures under large pose changes due to a lack of 3D understanding. Recently, several approaches~\cite{hong2022headnerf,khakhulin2022realistic,ma2023otavatar,li2023one,yu2023nofa,li2023generalizable} pursue animatable 3D head synthesis in order for better 3D consistency. They often resort to monocular 3DMM reconstructions~\cite{deng2019accurate,feng2021learning} to provide geometry or pose guidance due to a lack of large-scale multi-view training data. For example, \cite{li2023one, li2023generalizable} leverage a monocular face reconstructor~\cite{deng2019accurate} to predict 3DMM coefficients as guidance for learning 3D motions and canonical heads. \cite{khakhulin2022realistic} relies on reconstructed mesh from~\cite{feng2021learning} as initialization for head rigging. \cite{ma2023otavatar,yu2023nofa} combine 3DMM poses and expressions with a pre-trained 3D GAN~\cite{chan2021efficient} for reconstructing animatable head NeRFs. However, estimating 3DMMs and their poses from a monocular image can be inaccurate, thus limiting their performance in 3D head reenactment. \cite{siarohin2023unsupervised} learns 3D head geometries and deformation from monocular videos in an unsupervised manner, yet its synthesis quality lags further behind.

\paravspace
\paragraph{3D-aware portrait generation.} Recent studies~\cite{nguyen2019hologan,schwarz2020graf,chan2021pi,gu2021stylenerf,deng2022gram,chan2021efficient,chen2023mimic3d} demonstrate that by combining 3D representations such as NeRF with adversarial learning on monocular images, it is possible to learn a 3D-aware generator for multi-view image generation of an object category, especially human portrait. While earlier works~\cite{chan2021pi,chan2021efficient,deng2022gram,sun2022ide} focus on static portrait synthesis, recent methods~\cite{tang2022explicitly,wu2022anifacegan,bergman2022generative,sun2022controllable, sun2023next3d,xu2023omniavatar,wu2023aniportraitgan} introduce 3DMMs for extra animation control. For example, \cite{wu2022anifacegan,xu2023omniavatar} learn 3D deformation fields to mimic the motions of 3DMM for driving canonical NeRFs. \cite{sun2023next3d} rasterizes 3DMM meshes with neural textures onto the tri-planes for animatable NeRF generation. These 3D-aware GANs serve as powerful tools for monocular head avatar reconstruction when combined with advanced GAN inversion techniques~\cite{roich2021pivotal,xie2023high,yin2023nerfinvertor,deng2023learning,ko20233d}. Nevertheless, the inversion process also requires camera pose or 3DMM estimation from a single image which brings inaccuracy. In addition, weight correction~\cite{roich2021pivotal} of the pre-trained generator is often needed for out-of-domain images, which makes them less practical when applied to large-scale scenarios.

\paravspace
\paragraph{Learning with synthetic data.} Synthetic data are widely used for training deep learning models in various tasks~\cite{richter2016playing,qiu2017unrealcv,hodavn2019photorealistic,saito2019pifu,yao2020simulating,sitzmann2019scene,wood2021fake,kowalski2020config,yeh2022learning,wang2023rodin}. Typically for human portraits, traditional graphics pipelines are often used to create synthetic data for face analysis and synthesis. For instance, \cite{wood2021fake,wood22043d} leverage photorealistic raytracing renderer~\cite{CyclesRenderer} to render high-quality 3D face models for landmark detection. \cite{tuan2017regressing,zhu2017face,Reda_Zhu_2020_CVPR} fit 3DMMs from images and train a CNN to regress the coefficients for monocular 3D reconstruction. \cite{zhou2019deep} creates face images of various lightings via Spherical Harmonic (SH) model~\cite{basri2003lambertian} for portrait relighting. However, for photorealistic image synthesis, these CG-style data encounter large domain gaps and often require extra adaptations~\cite{deng2020disentangled,garbin2020high,yeh2022learning}. Compared to the CG data, 3D-aware GANs~\cite{chan2021efficient,sun2023next3d} can generate images with higher photorealism. A recent method~\cite{trevithick2023real} shows that images generated by a 3D-aware GAN can be used for learning one-shot novel view synthesis at high fidelity. Still, using synthetic data for one-shot 4D head synthesis is largely underexplored.

%--------------------------------------------------------------------

\section{Approach}

Given a source image $I_s$ and a driving image $I_d$, we learn a model to synthesize a 3D head with the appearance of $I_s$ and the motion of $I_d$, allowing free-view rendering. We adopt triplane-based NeRF~\cite{chan2021efficient} as the underlying 3D representation for its fidelity and efficiency. To effectively learn monocular NeRF reconstruction, we introduce a part-wise generative model~$\mathrm{G}$ (GenHead) to synthesize multi-view head images of diverse identities and motions as training data (Sec.~\ref{sec:genhead} and~\ref{sec:data}). With the large-scale synthetic data, we learn an animatable tri-plane reconstructor $\Psi$ to directly reconstruct 4D head NeRFs from monocular images via a feed-forward pass (Sec.~\ref{sec:oneshot}). A disentangled learning strategy is introduced to isolate the reconstruction and the reenactment process of $\Psi$ to improve its generalizability to real images (Sec.~\ref{sec:training}). Figure~\ref{fig:overview} shows an illustration of the overall framework. We describe each part in detail below.

\subsection{Part-wise Generative Head Model}\label{sec:genhead}
The goal of GenHead is to ``turn" accessible monocular data into 4D ones to enable learning the subsequent 4D NeRF reconstruction in a data-driven manner. To this end, we adopt the recent 3D-aware GAN framework~\cite{bergman2022generative, xu2023omniavatar, sun2023next3d} which effectively learns NeRF synthesis via adversarial learning on monocular images. Nonetheless, existing head GANs are not designed for full motion control of the face, eyes, mouth, and neck. Therefore, we introduce a part-wise generative model to deal with the complex head animation.

Specifically, the GenHead model $\mathrm{G}$ consists of a part-wise tri-plane generator $\mathrm{G}_{ca}$ for canonical head NeRF synthesis and a part-wise deformation field $\mathcal{D}$ for morphing the canonical head. $\mathrm{G}_{ca}$ receives a random noise $\bm z \in \mathbb{R}^{512}$ and a FLAME~\cite{li2017learning} shape code $\bm \alpha \in \mathbb{R}^{300}$, and generates two tri-planes for the head region as well as eyes and mouth:
\begin{equation}
    \mathrm{G}_{ca}: (\bm z, \bm \alpha) \rightarrow [T_{h}, T_{p}] \in \mathbb{R}^{256\times 256\times 96 \times 2}, \label{eq:genhead_ca}
\end{equation}
where $T_{h}$ and $T_{p}$ are tri-planes of a canonical 3D head and its eyes and mouth regions, respectively. Following~\cite{chan2021efficient}, we leverage a StyleGAN2~\cite{karras2020analyzing} backbone for $\mathrm{G}_{ca}$. A 3D point can be projected onto the tri-planes to obtain features $\bm f_h$ and $\bm f_p \in \mathbb{R}^{32}$ for radiance decoding and volume rendering (see \cite{chan2021efficient} for details). Note that the head synthesized by $\mathrm{G}_{ca}$ is aligned with FLAME's average face shape, and $\bm \alpha$ in Eq.~\eqref{eq:genhead_ca} is only for shape-related appearance but not shape variations, without which we found quality drops due to random combination of deformation and appearance.

The deformation field $\mathcal{D}$ receives the shape code $\bm \alpha$, together with FLAME's expression code $\bm \beta \in \mathbb{R}^{100}$ and pose code $\bm \gamma = [\bm \gamma_{eye}, \bm \gamma_{jaw}, \bm \gamma_{neck}] \in \mathbb{R}^{9}$, and outputs part-wise 3D deformations for a point $\bm x$ in the observation space:
 \begin{equation}
    \mathcal{D}: (\bm x, \bm \alpha, \bm \beta, \bm \gamma) \rightarrow [\Delta \bm x_{h}, \Delta \bm x_{p}] \in \mathbb{R}^{3\times2}.\label{eq:deform}
\end{equation}
By adding $\Delta \bm x_{h}$ or $\Delta \bm x_{p}$ to $\bm x$, it is deformed into the canonical spaces of $T_{h}$ or $T_{p}$ for feature acquisition, respectively. We utilize a FLAME mesh $\bm m(\bm \alpha,\bm \beta,\bm \gamma)$ to directly derive $\mathcal{D}$. Specifically, we first obtain deformation between $\bm m(\bm \alpha,\bm \beta,\bm \gamma)$ and an average face mesh $\bm m(\bm 0,\bm 0,\bm \gamma_{ca})$ for each vertex on $\bm m$, where $\bm \gamma_{ca}=[\bm 0, \bm \gamma_{jaw,ca},\bm 0]$ denotes a canonical pose with mouth open. Then, for a free-space point, we derive its deformation to the canonical spaces via a weighted average of that of its nearest vertices on $\bm m$. For $\Delta \bm x_h$, we do not consider the eye gaze $\bm \gamma_{eye}$. For $\Delta \bm x_p$, we adopt two different derivations around eyes and mouth: for the eye region, we use the deformation between $\bm m(\bm \alpha,\bm \beta,\bm \gamma)$ and $\bm m(\bm 0,\bm 0,\bm \gamma_{ca})$ as described above; for the mouth region, we use the deformation between $\bm m(\bm \alpha,\bm 0,\bm \gamma)$ and $\bm m(\bm 0,\bm 0,\bm \gamma_{ca})$ without expression $\bm \beta$, to handle relative movements between lips and teeth caused by expression. See Sec.~\ref{sec:genhead_supp} for more details and illustrations.

After obtaining features from each tri-plane, we perform volume rendering~\cite{kajiya1984ray,mildenhall2020nerf} to synthesize two feature maps from $\bm f_h$ and $\bm f_p$, respectively, and blend them via a rasterized mask of $\bm m(\bm \alpha,\bm \beta,\bm \gamma)$ at the corresponding viewpoint $\bm \theta$. We leverage another StyleGAN2 to generate 2D background $I_{bg}$, fuse it with the rendered foreground $I_{f}$, and send the result to a 2D super-resolution module for final image synthesis, as in~\cite{an2023panohead}. The whole model is trained end-to-end via adversarial loss~\cite{goodfellow2014generative} using a dual discriminator~\cite{sun2023next3d}. 

After training, GenHead is free to synthesize 4D head data by altering the combination of $(\bm z, \bm \alpha, \bm \beta, \bm \gamma, \bm \theta)$.

\subsection{4D Data Synthesis with GenHead}\label{sec:data}

We generate two types of data for learning the 4D head synthesis model, namely ``dynamic" data and ``static" data. The dynamic data are responsible for head reenactment which contain synthetic identities $(\bm z, \bm \alpha)$ with multiple motions $(\bm \beta, \bm \gamma)$ per subject and different camera poses $\bm \theta$ per subject and motion. The static data are used to enhance the 3D reconstruction generalizability, which contains only a single motion per identity with different camera views.

In particular, we extract $(\bm \alpha, \bm \beta, \bm \gamma, \bm \theta)$ from monocular images and videos via off-the-shelf 3DMM reconstruction methods~\cite{deng2019accurate,bfmtoflame} with further landmark-based optimization to form a sample set. During training the 4D head synthesizer, we randomly construct online pairs of $(\bm z, \bm \alpha)$ out of the sample set as the dynamic identities. For each identity, we assign random head motions and random camera poses per motion. 
Note that for each dynamic identity, we sample expression $\bm \beta$ extracted from the same video clip to alleviate the identity leakage issue born in the linear 3DMM model~\cite{jiang2019disentangled}. For the static data, we follow a similar procedure to construct random identities with an arbitrary motion per identity and random camera poses. 
We also keep intermediate outputs of GenHead as extra labels, including sampled tri-plane features $\bar{T}(\bm x)$, rendered low-resolution feature maps and backgrounds $\bar{I}_{f}$ and $\bar{I}_{bg}$, depth images $\bar{I}_{depth}$, and opacity images $\bar{I}_{opa}$, to facilitate learning in Sec.~\ref{sec:training}. More details and visualizations of the data are in Sec.~\ref{sec:data_supp}.

Notably, although we leverage monocular 3DMM reconstruction in building the synthetic data, we do not use the reconstructed 3DMM codes as input to the one-shot synthesis model as in~\cite{ma2023otavatar,yu2023nofa,li2023one}. This way, we avoid inheriting the 3DMM reconstruction errors at inference time.

\subsection{Animatable Triplane Reconstructor}\label{sec:oneshot}
\begin{figure*}[t]
	\small
	\centering
	\includegraphics[width=0.95\textwidth]{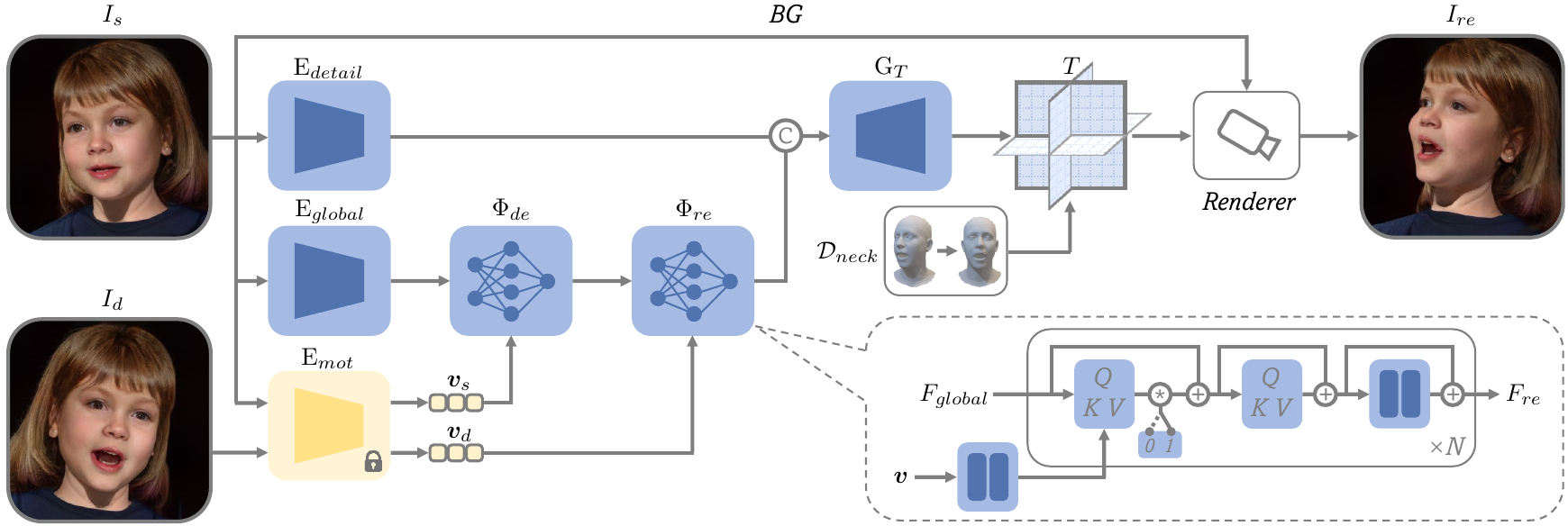}
	\vspace{-5pt}
	\caption{Architecture of the animatable triplane reconstructor $\Psi$. An encoder $\mathrm{E}_{global}$ first extracts the appearance feature map of $I_s$. The feature is then sent to a canonicalization and reenactment module $\Phi$ consisting of a de-expression module $\Phi_{de}$ and a reenactment module $\Phi_{re}$ sharing the same structure, which receives motion features from either $I_s$ or $I_d$ for expression neutralization or motion injection accordingly. The reenacted feature is then concatenated with a detail feature map from another encoder $\mathrm{E}_{detail}$, and sent to a decoder $\mathrm{G}_{T}$ to synthesize a tri-plane $T$, bearing the appearance of $I_s$ and the motion of $I_d$. With a FLAME-derived 3D deformation field $\mathcal{D}_{neck}$ to handle neck pose and a volumetric renderer with 2D super-resolution, $T$ can be rendered to a reenacted image $I_{re}$ at an arbitrary view.} 
	\label{fig:framework}
\vspace{-5pt}
\end{figure*}

Our 4D head synthesis model builds upon an animatable tri-plane reconstructor $\Psi$, which takes a source image $I_s$ and a driving image $I_d$ as input, and reconstructs a tri-plane $T \in \mathbb{R}^{256\times256\times96}$ for synthesizing reenacted image $I_{re}$ at a camera pose $\bm \theta$ with a renderer $\mathcal{R}$:
 \begin{equation}
 \Psi: (I_s,I_d) \rightarrow  T,\; \mathcal{R}: (T, \bm \theta) \rightarrow I_{re},
    \label{eq:reconstructor}
\end{equation}
where $\mathcal{R}$ consists of radiance decoding, volume rendering, and 2D super-resolution, similar to Sec.~\ref{sec:genhead}. The tri-plane $T$ in Eq.~\eqref{eq:reconstructor} represents a 3D head with the identity of $I_s$ and the motion of $I_d$ (with zero neck pose) instead of an average head as in GenHead. To faithfully obtain $T$, we construct $\Psi$ with two appearance encoders $\mathrm{E}_{global}$ and $\mathrm{E}_{detail}$, a motion encoder $\mathrm{E}_{mot}$, a canonicalization and reenactment module $\Phi$, and a tri-plane decoder $\mathrm{G}_{T}$, as shown in Fig.~\ref{fig:framework}. 

$\mathrm{E}_{global}$ and $\mathrm{E}_{detail}$ are used to extract appearance feature maps from $I_s$ for the subsequent 3D reconstruction and reenactment, which adopt the CNN structures in~\cite{trevithick2023real}.

Then, we introduce $\Phi$ to canonicalize the feature map $F_{global}$ from $\mathrm{E}_{global}$ as well as reenact it with the motion from $I_d$. $\Phi$ consists of two transformer-based modules $\Phi_{de}$ and  $\Phi_{re}$ sharing the same structure, one for neutralizing the expression of $F_{global}$ and the other for injecting motions of $I_d$ for reenactment, meanwhile they both serve for 3D pose canonicalization. Specifically, they contain multiple transformer blocks each with a cross-attention layer, a self-attention layer, and an MLP. The self-attention and the MLP layers are for pose canonicalization. The cross-attention layers receive additional motion features of either $I_s$ or $I_d$, and add them onto the global feature map of $I_s$ for de-expression or reenactment accordingly, where the global feature map provides queries while the motion features provide keys and values, as depicted in Fig.~\ref{fig:framework}. Notably, introducing $\Phi_{de}$ for expression neutralization is important. Otherwise, we find the self-attention and the MLP layers take over the de-expression job which easily leads to overfitting. What's more, this architecture enables us to disentangle the learning of 3D reconstruction and reenactment, by simply multiplying the output of all cross-attention layers by zeros or not. This strategy is crucial to the model's generalizability on real images and will be further discussed in Sec.~\ref{sec:training}.

Besides, the motion features provided to $\Phi$ are equally important as they are required to capture identity-irrelevant animations. To this end, we utilize a pre-trained motion encoder~\cite{wang2023progressive} as $\mathrm{E}_{mot}$, which is learned via reconstructing large-scale monocular videos for faithful cross-identity reenactment. $\mathrm{E}_{mot}$ predicts a motion vector $\bm v \in \mathbb{R}^{548}$ from an image, which is then sent to $\Phi$ to compute the cross-attentions. Compared to FLAME's expression $\bm \beta$, $\bm v$ is more identity-agnostic, yielding better results when combined with our 4D synthetic data, as we will show in Sec.~\ref{sec:ablation}.

Finally, the reenacted feature map $F_{re}$ after $\Phi$ is concatenated with the detail feature map $F_{detail}$ from $\mathrm{E}_{detail}$, and sent into a decoder $\mathrm{G}_{T}$~\cite{trevithick2023real} to obtain the tri-plane $T$. Then, $T$ is rendered to the reenacted image $I_{re}$ via Eq.~\eqref{eq:reconstructor}. Before the volume rendering process, we apply a FLAME-derived deformation field $\mathcal{D}_{neck}$ to handle the neck pose rotation similarly as in Sec.~\ref{sec:genhead}. Since the neck pose change yields almost uniform rigid transformations, this strategy is acceptable as it does not require highly accurate 3DMM reconstructions (see Sec.~\ref{sec:ablation}). Besides, we use a shallow U-Net to predict a 2D background feature map $I_{bg}$ and blend it with the foreground $I_{f}$ for rendering. 

The whole reconstructor $\Psi$, except the fixed motion encoder $\mathrm{E}_{mot}$, is trained end-to-end using the synthetic data from GenHead, as described below.

\subsection{Disentangled Learning via Synthetic 4D Data}\label{sec:training}
We learn the synthesis model $\Psi$ in a self-reenacting manner, where we randomly choose two images of the same identity as $I_s$ and $I_d$, respectively, and choose another image with $I_d$'s identity and motion but a different camera view as the ground truth reenacted image $\bar{I}_{re}$. During training, $\Psi$ is enforced to synthesize an image $I_{re}$ via Eq.~\eqref{eq:reconstructor} to match the content of $\bar{I}_{re}$ at its viewing point, via a serial of losses which will be described later. 

While the training process relies solely on synthetic data, we introduce a disentangled learning strategy to generalize the learned model to real data. Specifically, we find that the model's generalizability highly depends on the function of all non-cross-attention layers in $\Phi$. By default, $\Phi$ is inclined to use the self-attention and MLP layers in $\Phi_{de}$ to handle both the expression neutralization and pose canonicalization of $I_s$ (see Fig.~\ref{fig:deexp} for an illustration), causing the network to quickly overfit to the synthetic identities. 

\begin{figure*}[t]
	\small
	\centering
	\includegraphics[width=1.0\textwidth]{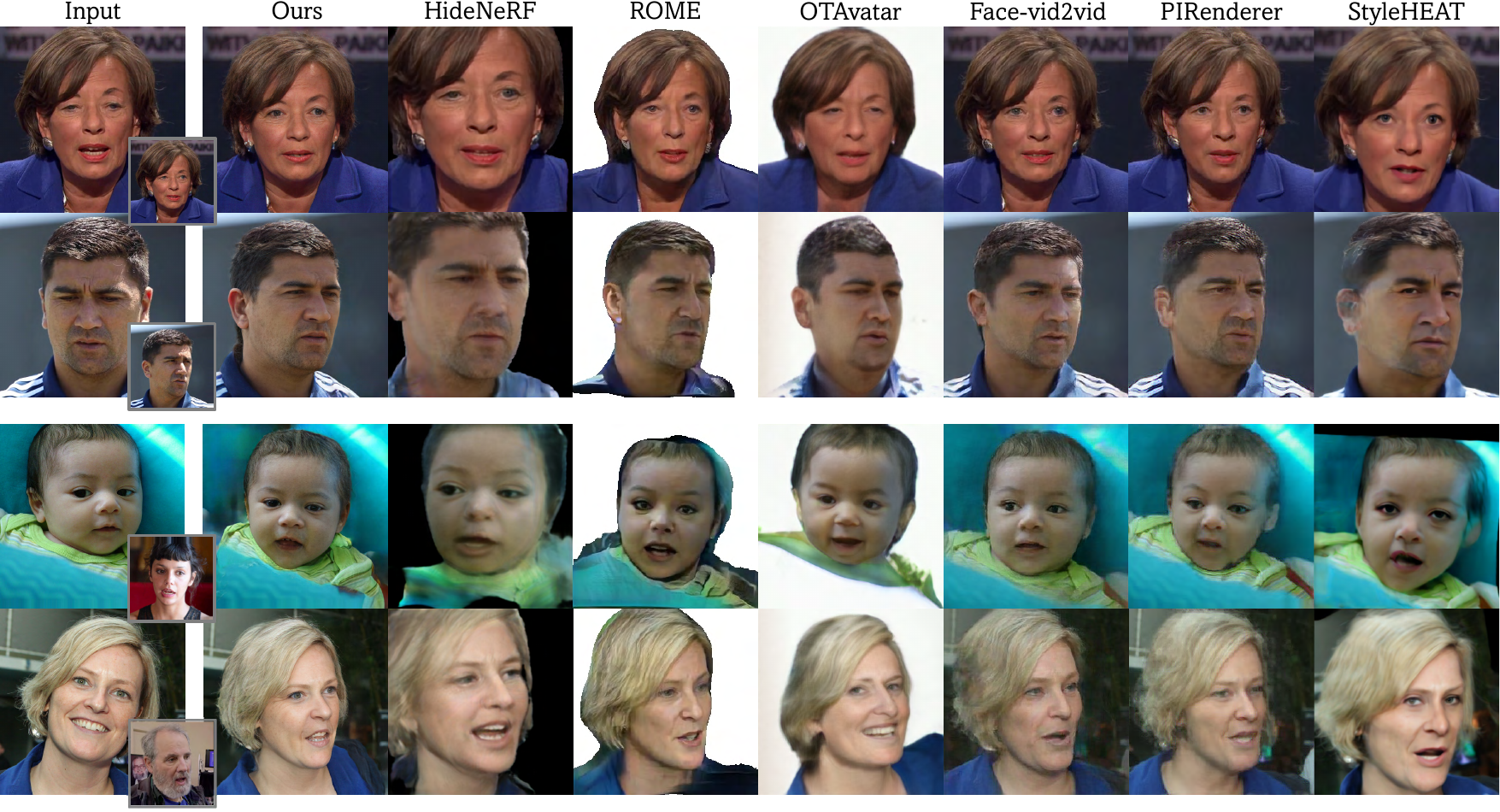}
	\vspace{-15pt}
	\caption{Qualitative comparison on one-shot head reenactment with previous methods. \textbf{Best viewed with zoom-in.}} 
	\label{fig:compare}
\vspace{-5pt}
\end{figure*}

To tackle this problem, our intuition is to disentangle the self-attentions and MLPs to focus on pose canonicalization only and let the cross-attention layers to deal with all the motion-related processes. We achieve this by simply multiplying zeros by the outputs of all the cross-attention layers at a fixed probability, and using the static data described in Sec.~\ref{sec:data} to execute the self-reenacting process at the same time. This way, the training process randomly degenerates to a static 3D reconstruction process for the remaining layers in $\Phi$, meanwhile, all the cross-attention layers will be trained normally in a default reenactment process leveraging the dynamic data. Empirically, this strategy largely improves the reconstruction fidelity on real images (Sec.~\ref{sec:ablation}).

The overall training objective is as follows:
\begin{equation}
\vspace{-1pt}
\mathcal{L} = \mathcal{L}_{re} + \mathcal{L}_{f} + \mathcal{L}_{tri} + \mathcal{L}_{depth} + \mathcal{L}_{opa} + \mathcal{L}_{id} + \mathcal{L}_{adv}, \label{eq:training}
\end{equation}
where $\mathcal{L}_{re}$ calculates the perceptual difference~\cite{zhang2018unreasonable} and L1 distance between $I_{re}$ and its ground truth $\bar{I}_{re}$. $\mathcal{L}_{f}$ calculates the L1 distance between $I_{f},I_{bg}$ and their corresponding ground truth. $\mathcal{L}_{tri}$ computes the L1 difference between sampled tri-plane features $T(\bm x)$ and $\bar{T}(\bm x)$. Note that we do not calculate the difference between $T$ and $[T_h,T_p]$ in Eq.~\eqref{eq:genhead_ca} directly as in~\cite{trevithick2023real} because the tri-planes represent different geometries in $\Psi $ and in GenHead. $\mathcal{L}_{depth}$ computes the L1 difference between $I_{depth}$ and $\bar{I}_{depth}$. $\mathcal{L}_{opa}$ is the L1 difference between $I_{opa}$ and $\bar{I}_{opa}$. $\mathcal{L}_{id}$ calculates the negative cosine similarity between face recognition features~\cite{deng2019arcface} of $I_{re}$ and $\bar{I}_{re}$. $\mathcal{L}_{adv}$ is the adversarial loss between $I_{re}$ and $\bar{I}_{re}$ leveraging the discriminator of GenHead. 

%--------------------------------------------------------------------

\section{Experiments}\label{sec:experiment}

\paragraph{Implementation details.} We train GenHead with a re-aligned version of FFHQ~\cite{karras2019style} at $512^2$ resolution. For 4D data synthesis, we use 3DMM parameters extracted from FFHQ and VFHQ~\cite{xie2022vfhq}. The camera poses are sampled from a pre-defined distribution. See~\ref{sec:implement} for more details.

\paravspace
\paragraph{Evaluation of GenHead.} We leave the evaluation of GenHead in~Sec.~\ref{sec:supp_results_genhead}.

\begin{table*}[t]\footnotesize
\centering
\caption{Self-reenactment results on VFHQ at $512^2$. LPIPS$_h$ is for head region. \tikzcircle[gold,fill=gold]{2pt}, \tikzcircle[silver,fill=silver]{2pt}, and \tikzcircle[bronze,fill=bronze]{2pt} denote the 1st, 2nd and 3rd places, respectively. }
\vspace{-5pt}
\begin{tabular}{l|lll|lll|lll|lll}
\toprule

\multicolumn{1}{c}{\multirow{2}{*}{Method}} &  \multicolumn{3}{c}{All}  & \multicolumn{3}{c}{Yaw Diff. $<15^{\circ}$} & \multicolumn{3}{c}{Yaw Diff. $15^{\circ}\sim30^{\circ}$} & \multicolumn{3}{c}{Yaw Diff. $>30^{\circ}$}\\ 
\cline{2-13} 
 \multicolumn{1}{c}{} & \!\!LPIPS$_{h}$$\downarrow$\!\!\! & LPIPS $\downarrow$ &  FID $\downarrow$ & ID $\uparrow$   & AED $\downarrow$   & APD $\downarrow$   & ID $\uparrow$ &  AED $\downarrow$ &  APD $\downarrow$  & ID $\uparrow$ & AED $\downarrow$ & APD $\downarrow$\\
 \midrule
 PIRenderer~\cite{ren2021pirenderer} & 0.214\tikzcircle[bronze,fill=bronze]{2pt} & 0.325 &  61.0\tikzcircle[bronze,fill=bronze]{2pt} & 0.820\tikzcircle[silver,fill=silver]{2pt}  & 0.031\tikzcircle[bronze,fill=bronze]{2pt}  & 1.245 & 0.647\tikzcircle[bronze,fill=bronze]{2pt} & 0.052 & 2.483\tikzcircle[bronze,fill=bronze]{2pt} & 0.536\tikzcircle[silver,fill=silver]{2pt} & 0.077\tikzcircle[bronze,fill=bronze]{2pt} & 9.326\\ 
Face-vid2vid~\cite{wang2021one} & 0.187\tikzcircle[silver,fill=silver]{2pt} & 0.290\tikzcircle[gold,fill=gold]{2pt} & 52.6\tikzcircle[silver,fill=silver]{2pt} & 0.848\tikzcircle[gold,fill=gold]{2pt} & 0.025\tikzcircle[gold,fill=gold]{2pt}  & 0.778\tikzcircle[silver,fill=silver]{2pt} & 0.665\tikzcircle[silver,fill=silver]{2pt} & 0.048\tikzcircle[silver,fill=silver]{2pt} & 4.035 & 0.502 & 0.098 & 24.99 \\
StyleHEAT~\cite{yin2022styleheat} & 0.224 & 0.321\tikzcircle[bronze,fill=bronze]{2pt} & 67.3 & 0.649  & 0.050  &  1.392 & 0.524 & 0.068 & 2.727 & 0.420 & 0.106 & 10.89 \\
\midrule
ROME~\cite{khakhulin2022realistic} & 0.327 & 0.598 & 110 & 0.671 & 0.034 & 0.960\tikzcircle[bronze,fill=bronze]{2pt} & 0.589  & 0.046\tikzcircle[gold,fill=gold]{2pt} & 1.279\tikzcircle[silver,fill=silver]{2pt} & 0.497 & 0.076\tikzcircle[silver,fill=silver]{2pt} & 1.671\tikzcircle[silver,fill=silver]{2pt}\\
OTAvatar~\cite{ma2023otavatar} & 0.291 & 0.557 & 112  &  0.448 & 0.077 & 3.306 & 0.339 &0.089&7.396& 0.251 & 0.177& 8.157 \\
HideNeRF~\cite{li2023one} & 0.281  & 0.418 & 80.5 & 0.820\tikzcircle[bronze,fill=bronze]{2pt} & 0.032 & 1.762 & 0.642 &0.060&6.001&0.516\tikzcircle[bronze,fill=bronze]{2pt}&0.084&7.879\tikzcircle[bronze,fill=bronze]{2pt} \\
Ours& 0.181\tikzcircle[gold,fill=gold]{2pt} & 0.320\tikzcircle[silver,fill=silver]{2pt} & 43.0\tikzcircle[gold,fill=gold]{2pt} & 0.790 & 0.030\tikzcircle[silver,fill=silver]{2pt} & 0.560\tikzcircle[gold,fill=gold]{2pt} & 0.668\tikzcircle[gold,fill=gold]{2pt} &0.051\tikzcircle[bronze,fill=bronze]{2pt}&0.811\tikzcircle[gold,fill=gold]{2pt} & 0.580\tikzcircle[gold,fill=gold]{2pt} & 0.074\tikzcircle[gold,fill=gold]{2pt} & 1.340\tikzcircle[gold,fill=gold]{2pt}\\
\bottomrule
\end{tabular}

\label{tab:main_self}
\vspace{-5pt}
\end{table*}

\begin{table}[t]\footnotesize
\centering
\caption{Cross-reenactment results on VFHQ dataset at $512^2$.}
\vspace{-5pt}
\begin{tabular}{l|llll}
\toprule

Method  &  FID $\downarrow$ &  ID $\uparrow$  & AED $\downarrow$ & APD $\downarrow$\\
 \midrule
 PIRenderer~\cite{ren2021pirenderer} &  78.0\tikzcircle[silver,fill=silver]{2pt} & 0.582\tikzcircle[bronze,fill=bronze]{2pt} & 0.146\tikzcircle[gold,fill=gold]{2pt} & 3.241\tikzcircle[bronze,fill=bronze]{2pt}\\ 
Face-vid2vid~\cite{wang2021one} &  78.6\tikzcircle[bronze,fill=bronze]{2pt} & 0.606\tikzcircle[silver,fill=silver]{2pt} & 0.179 & 7.771 \\
StyleHEAT~\cite{yin2022styleheat} &  81.1 & 0.499 & 0.165 & 5.932 \\
\midrule
ROME~\cite{khakhulin2022realistic} & 107 & 0.532 & 0.148\tikzcircle[silver,fill=silver]{2pt} & 2.936\tikzcircle[silver,fill=silver]{2pt}\\
OTAvatar~\cite{ma2023otavatar} & 107&0.350& 0.194 & 6.509 \\
HideNeRF~\cite{li2023one} &104&0.491&0.165&4.208 \\
Ours& 54.8\tikzcircle[gold,fill=gold]{2pt} &0.620\tikzcircle[gold,fill=gold]{2pt} & 0.164\tikzcircle[bronze,fill=bronze]{2pt} & 1.020\tikzcircle[gold,fill=gold]{2pt}\\
\bottomrule
\end{tabular}

\label{tab:main_cross}
\vspace{-5pt}
\end{table}

\subsection{One-Shot 4D Head Synthesis Results}
Figure~\ref{fig:teaser} shows our one-shot head synthesis results on real images in FFHQ (more in Sec.~\ref{sec:results_main_supp}). Our method faithfully reconstructs a 3D head with rich details from a monocular image. The proxy 3D shapes extracted via Marching Cubes~\cite{lorensen1987marching} depict reasonable 3D geometry recoveries, which is the key for 3D-consistent reenactment. We can freely change the neck pose of the reconstructed 3D heads, as well as varying their expressions given different driving images. We also support novel view synthesis thanks to the underlying 3D representation. Besides, backgrounds can be well separated during foreground animations. Our inference pipeline runs at $10$ FPS on a single A100 GPU.

\subsection{Comparison with Prior Arts}
\paragraph{Baselines.} We compare our method with existing one-shot head avatar synthesis methods, including 2D-based ones: PIRenderer~\cite{ren2021pirenderer}, Face-vid2vid~\cite{wang2021one}, and StyleHEAT~\cite{yin2022styleheat}; and 3D-based ones: ROME~\cite{khakhulin2022realistic}, OTAvatar~\cite{ma2023otavatar}, and HideNeRF~\cite{li2023one}.

\paravspace
\paragraph{Metrics.} We conduct self-reenactment and cross-identity reenactment on $100$ test video clips in VFHQ. We use LPIPS and Fr\'echet Inception Distances (FID)~\cite{heusel2017gans} to measure the image synthesis quality, ID to measure the identity similarity~\cite{deng2019arcface} between the reenacted results and the appearance images, Average Expression Distance (AED)~\cite{lin20223d} for expression control accuracy, and Average Pose Distance (ADP, $\times 1000$ by default)~\cite{chan2021efficient} for pose control accuracy. 

\paravspace
\paragraph{Qualitative results.} Figure~\ref{fig:compare} shows visual comparisons on samples from VFHQ and FFHQ. Our method synthesizes head images with higher fidelity and definition compared to the alternatives. What's more, our method consistently yields better results compared to the others when the head pose differences between the sources and the drivings are significant, where we largely preserve the identities and head shapes. By contrast, the 2D-based methods experience large shape distortions. In addition, although HideNeRF, ROME, and OTAvatar also adopt 3D representations, their learning on monocular real data with imperfect 3D estimation impairs the geometry accuracy. More results can be found in Sec.~\ref{sec:compare_supp}.   

\paravspace
\paragraph{Quantitative results.} Table~\ref{tab:main_self} and~\ref{tab:main_cross}  show the quantitative comparisons on self-reenactment and cross-reenactment, respectively. In the self-reenactment, we divide the source and driving pairs into three groups based on their yaw angle differences. From Tab.~\ref{tab:main_self}, our reconstruction fidelity largely outperforms other 3D-based methods, and even surpasses two 2D-based approaches: PIRenderer and StyleHEAT. We also obtain comparable results with Face-vid2vid. Besides, our method yields the best pose accuracy, and achieves the highest identity similarities under large pose variations. This is inline with the visual results that our reconstructed geometries are more reasonable. We also demonstrate competitive expression control ability in terms of AED, whereas we solely learn on synthetic data without seeing real videos.

\begin{figure}[t]
\vspace{-2pt}
	\small
	\centering
	\includegraphics[width=1.0\columnwidth]{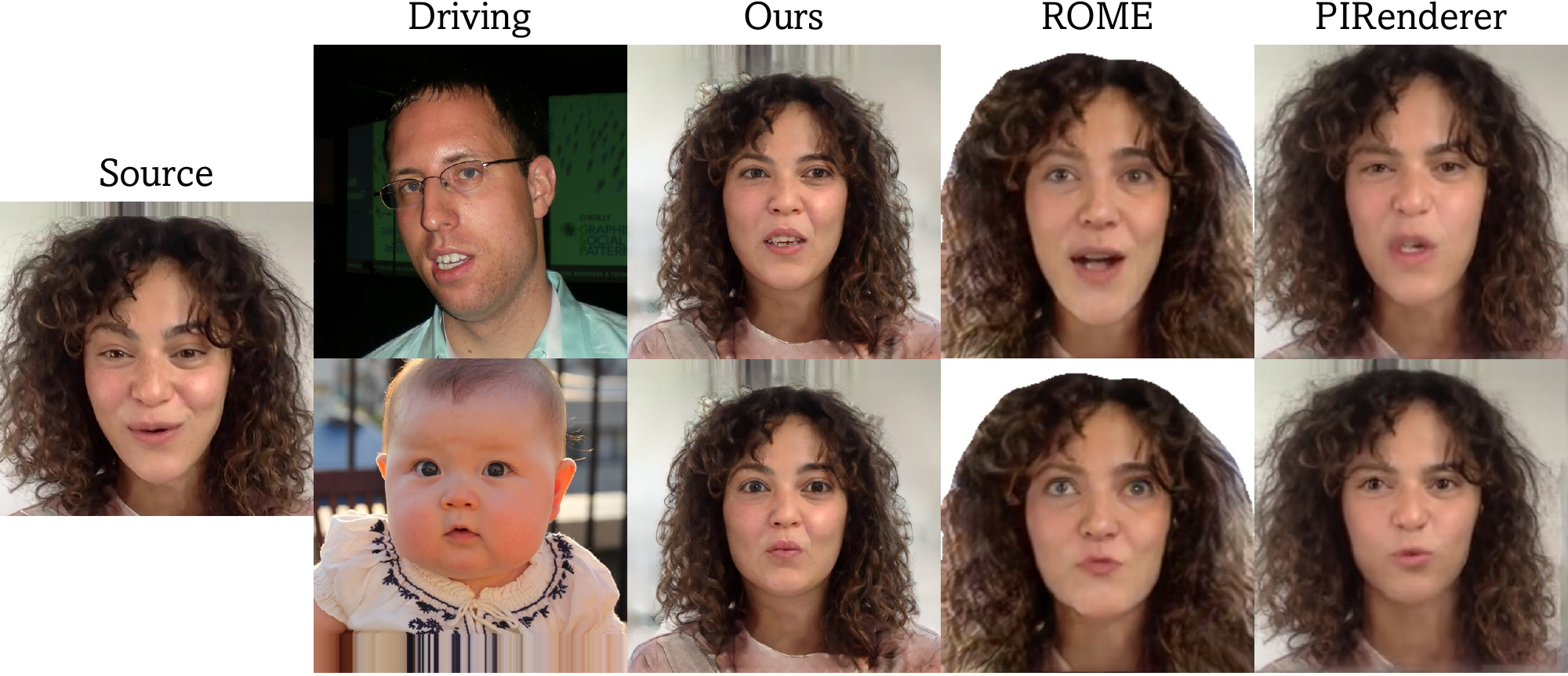}
	\vspace{-10pt}
	\caption{Reenactment results with different driving targets.} 
	\label{fig:compare_shape}
  \vspace{-5pt}
\end{figure}

In the cross-reenactment in Tab.~\ref{tab:main_cross}, we obtain the best image quality, identity similarity, and pose accuracy, with competitive expression accuracy. Note that although PIRenderer and ROME achieve lower AED, they tend to stretch the source geometry to match the absolute shapes of the drivings, leading to undesired identity changes, as shown in Fig.~\ref{fig:compare_shape}. This is a known identity leakage issue of the linear 3DMMs~\cite{paysan20093d,li2017learning}, and can influence the AED metric which is also based on 3DMM coefficients. By contrast, our method is more robust to shape variations of the drivings, thanks to the identity-agnostic motion feature of~\cite{wang2023progressive}.

\subsection{Ablation Study}\label{sec:ablation}

We conduct ablation studies on self-reconstruction and cross-reenactment using the first $1$K identities in FFHQ to validate the efficacy of different components. \textit{\textbf{A}} is a baseline that removes the cross-attention layers in $\Phi_{de}$, the disentangled learning strategy, and the static data. \textit{\textbf{B}} adds the learning strategy back by randomly selecting $50\%$ identities in the dynamic data for static 3D reconstruction. \textit{\textbf{C}} uses the complete structure of $\Phi_{de}$. \textit{\textbf{D}} combines \textit{B} and \textit{C} with both the above two components. \textit{\textbf{E}} is our final configuration, with extra static data for the 3D reconstruction process during disentangled learning. \textit{\textbf{F}} replaces the motion features of~\cite{wang2023progressive} with the FLAME codes $[\bm \beta, \bm \gamma_{eye},\bm \gamma_{jaw}]$. \textit{\textbf{G}} uses opacity images extracted via~\cite{cheng2021per} instead of the rendered ones from GenHead for $\mathcal{L}_{opa}$. \textit{\textbf{H}} utilizes monocular real data in FFHQ and VFHQ for training instead of the multi-view synthetic data. For efficiency, all configurations are trained to see $5000$K images in total. 

\begin{table}[t]
\vspace{-2pt}
    \centering
    \footnotesize
    \caption{Ablation study of our framework on FFHQ.}       \label{tab:ablation}
    \vspace{-1pt}
    \begin{tabular}{l|l|cccc}
    \toprule[1pt]
    \multicolumn{2}{c}{\multirow{2}{*}{Configurations}} & \multicolumn{2}{c}{Recon.}& \multicolumn{2}{c}{Cross-Reenact.}\\
    \cline{3-6} 
    \multicolumn{1}{c}{} &\multicolumn{1}{c}{} &\! \!LPIPS$_h$ $\downarrow$\!\! & \!\!ID $\uparrow$\! \!&\! \!AED $\downarrow$\!\!& \!\!APD $\downarrow$\!\! \\
    \midrule
    \!A\!\! &\!\!Baseline & 0.244 & 0.239 &0.179 & 0.963\\
  \!B\!\!&\!\!+ Disen. learning & 0.238& 0.281 &0.182 & 0.994\\
  \!C\!\!& \!\!+ $\Phi_{de}$   & 0.247 & 0.227 & 0.179 & 0.963\\
   \!D\!\! &\!\!B + C & 0.212 & 0.441 & 0.183 & 0.947\\
   \midrule
    \!E\!\!& \!\!D + Static (Ours)\!\! &0.188 & 0.703 & 0.193 & 1.022\\
    \midrule 
    \!F\!\!& \!\!with FLAME mot & 0.213 & 0.451 & 0.150 & 0.879\\
    \!G\!\!& \!\!with Detected $\bar{I}_{opa}$ & 0.198 & 0.552 & 0.189 & 1.080 \\
    \!H\!\!& \!\!with Real data & 0.123 & 0.960 & 0.263 & 7.537 \\

    \bottomrule[1pt]
    \end{tabular}
\vspace{-3pt}
\end{table}

\begin{figure}[t]
	\small
	\centering
	\includegraphics[width=1.0\columnwidth]{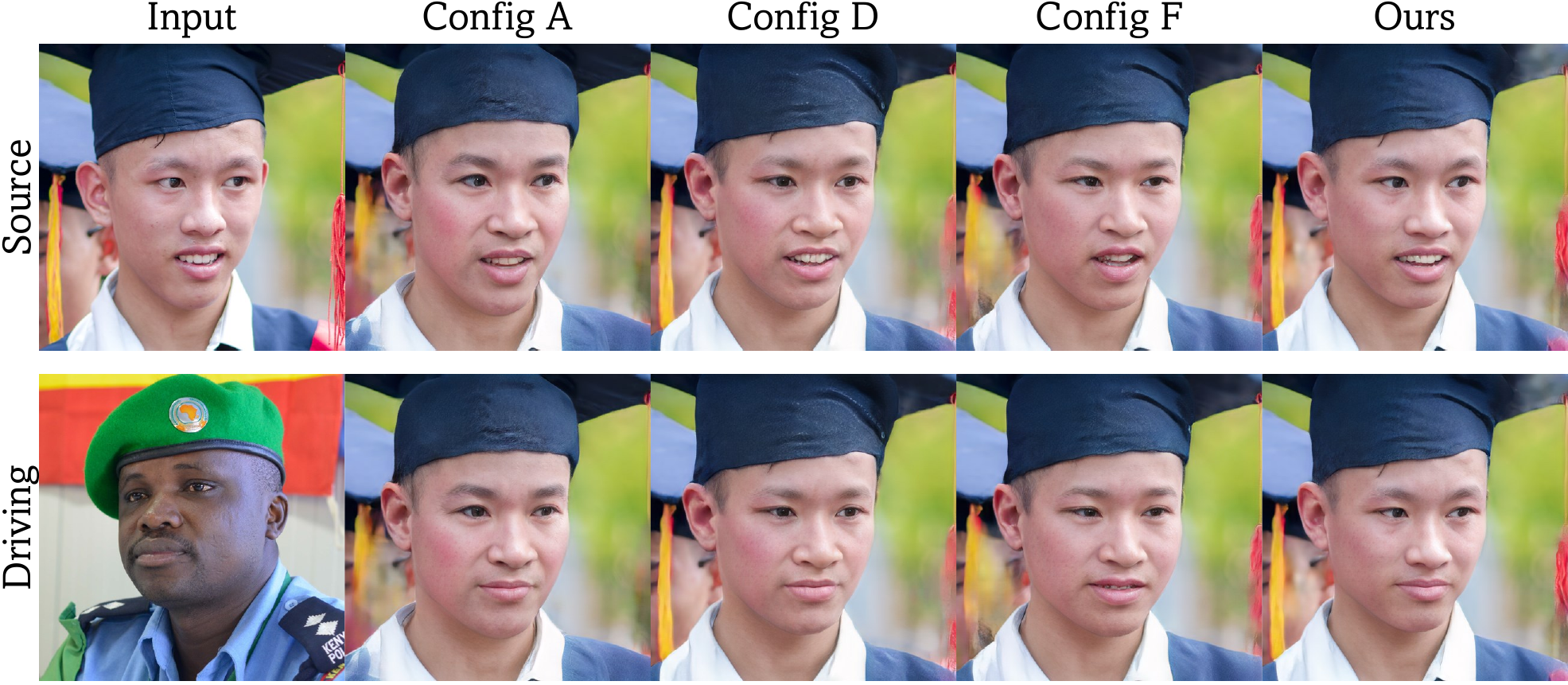}
	\vspace{-15pt}
	\caption{Reconstruction and driving results of different settings.} 
	\label{fig:ablation}
 \vspace{-3pt}
\end{figure}

\begin{figure}[t]
	\small
	\centering
	\includegraphics[width=1.0\columnwidth]{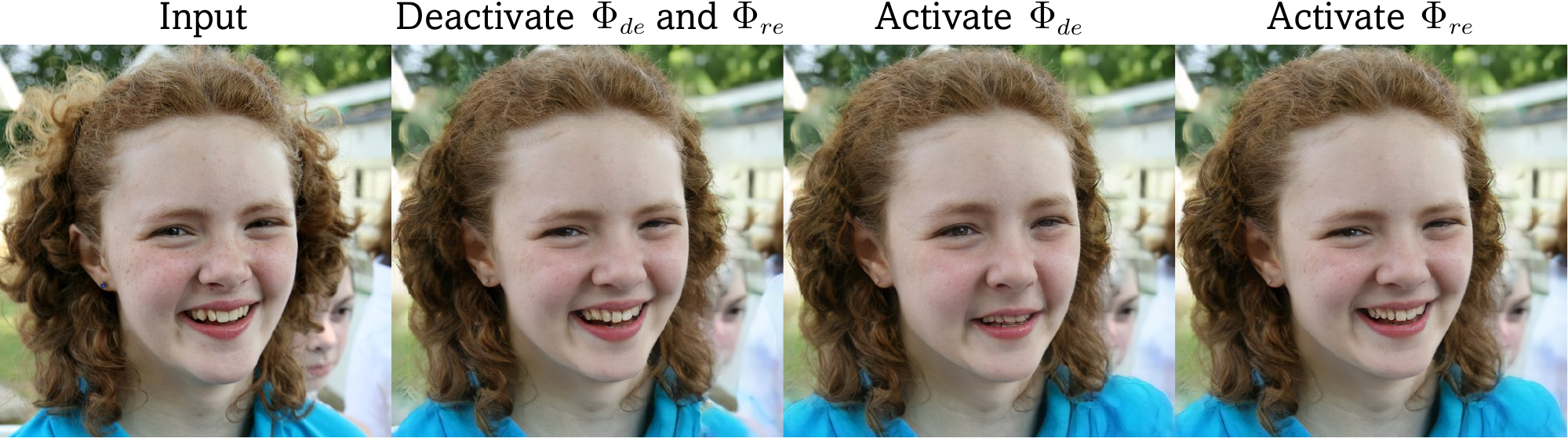}
	\vspace{-15pt}
	\caption{Function of learned $\Phi_{de}$ and $\Phi_{re}$.} 
	\label{fig:ablation_component}
  \vspace{-3pt}
\end{figure}

\paravspace
\paragraph{Module structure and learning strategy.} From Tab.~\ref{tab:ablation} and Fig.~\ref{fig:ablation}, the baseline setting leads to poor identity reconstruction results. Directly adding the disentangled learning strategy on top of the baseline slightly improves the reconstruction fidelity. Further introducing the de-expression module ensures a better disentanglement between 3D reconstruction and reenactment and effectively improves model's generalizability to real images. Finally, adding static data with more diverse identities for the 3D reconstruction process yields further improvement, with only minor influence to the reenactment accuracy. Figure~\ref{fig:ablation_component} illustrates the function of the learned $\Phi_{de}$, which turns the face into a canonical expression with mouth slightly open. Further sending the source motion to the reenactment module $\Phi_{re}$ after the de-expression step can faithfully recover the input image.

\paravspace
\paragraph{Influence of motion feature.} As shown in Tab.~\ref{tab:ablation}, using the FLAME codes as motion feature leads to better AED and APD, yet the reconstruction fidelity largely decreases. This alternative encounters a similar issue as other 3DMM-based methods in Fig.~\ref{fig:compare_shape}. If the driving image has very different face features with those of the source, this setting can lead to a semantic change of the given expression or even identity change, as shown in Fig.~\ref{fig:ablation}. We conjecture that such identity leakage issue confuses the model during training, leading to inferior generalizability for reconstruction.

\begin{figure}[t]
	\small
	\centering
	\includegraphics[width=1.0\columnwidth]{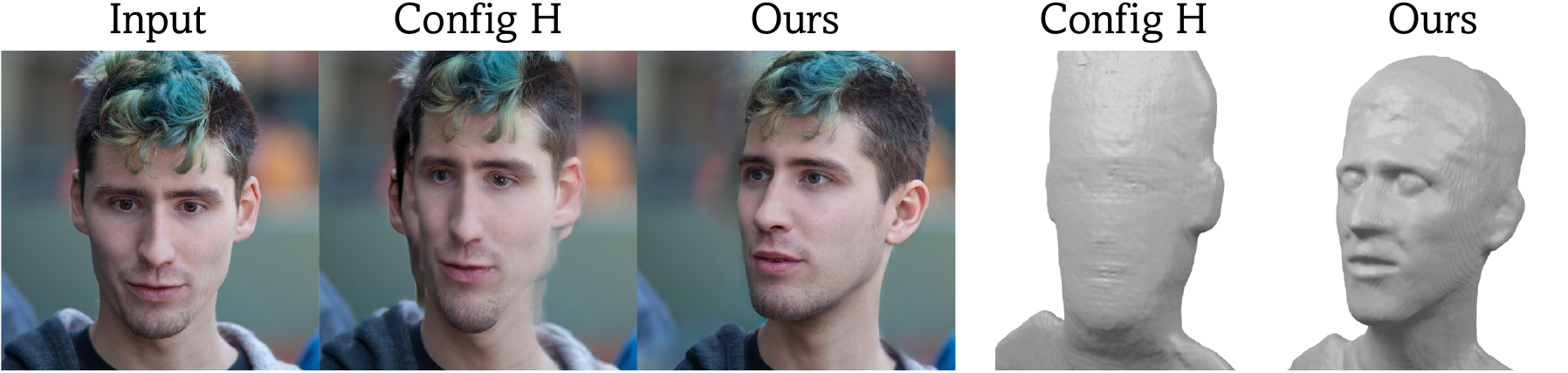}
	\vspace{-15pt}
	\caption{Comparison between models learned on different data.} 
	\label{fig:ablation_real}
\end{figure}

\begin{figure}[t]
	\small
	\centering
	\includegraphics[width=1.0\columnwidth]{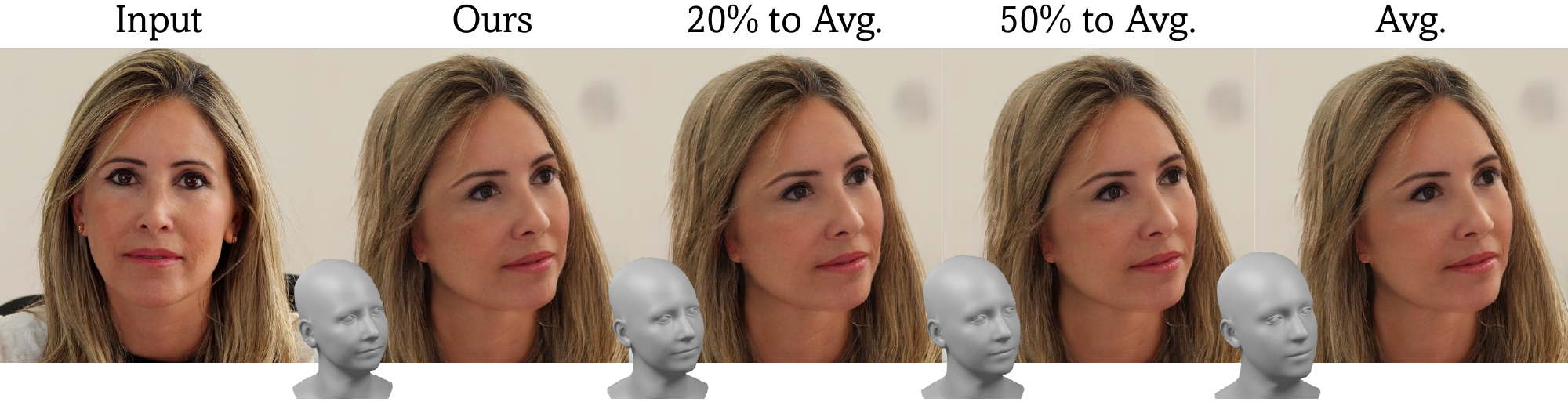}
	\vspace{-15pt}
	\caption{Neck pose control given different FLAME shapes.} 
	\label{fig:ablation_headpose}
 \vspace{-3pt}
\end{figure}

\paravspace
\paragraph{Synthetic data \textit{vs.}\ real data.} From Tab.~\ref{tab:ablation}, using detected opacity images is detrimental to the reconstruction fidelity. We conjecture that these detected results are not 3D-consistent across different views compared to the rendered synthetic ones, and thus increase the burden of the model to learn reasonable 3D reconstruction. What's more, if we replace the mult-view synthetic data with the monocular real ones, the learned geometry dramatically degenerates as indicated by the large APD and as shown in Fig.~\ref{fig:ablation_real}. This validates the importance of multi-view data with precise labels for learning reasonable 3D geometries.

\paravspace
\paragraph{Robustness of neck pose.} Figure~\ref{fig:ablation_headpose} demonstrates the robustness of our neck pose control given different FLAME shapes to derive $\mathcal{D}_{neck}$. We gradually deform an accurate FLAME mesh to an average shape. As shown, this process has negligible impact to the final synthesis quality.

%----------------------------------------------------------------

\section{Conclusions}
We presented a framework for learning high-fidelity one-shot 4D head synthesis via large-scale synthetic data. The core idea is to first learn a 4D generative model on monocular images via adversarial learning for multi-view data synthesis of diverse identities and full motion control; then utilize a transformer-based animatable tri-plane reconstructor to learn 4D head reconstruction via the synthetic data. A disentangled learning strategy is also introduced to enhance model's generalizability to real images. Experiments have demonstrated our superiority over previous works, and revealed our potential for large-scale head avatar creation.

\paravspace
\paragraph{Limitations and ethics consideration.} We leave the discussions in Sec.~\ref{sec:discuss}.

\section*{Acknowledgements}
We thank Zixin Yin and Deyu Zhou for helping with the early exploration of GenHead.

{
    \small
    \bibliographystyle{ieeenat_fullname}
    \bibliography{main}
}

\clearpage
\appendix

\twocolumn[
\centering
\Large
\textbf{Portrait4D: Learning One-Shot 4D Head Avatar Synthesis using Synthetic Data \\—— Appendix ——}
\vspace{0.5em}
]

\renewcommand{\thesection}{\Alph{section}}
\renewcommand{\thefigure}{\Roman{figure}}
\renewcommand{\thetable}{\Roman{table}}
\renewcommand{\theequation}{\Roman{equation}}
\setcounter{section}{0}
\setcounter{figure}{0}
\setcounter{equation}{0}

\begin{strip}
	\centering
	\includegraphics[width=1.0\textwidth]{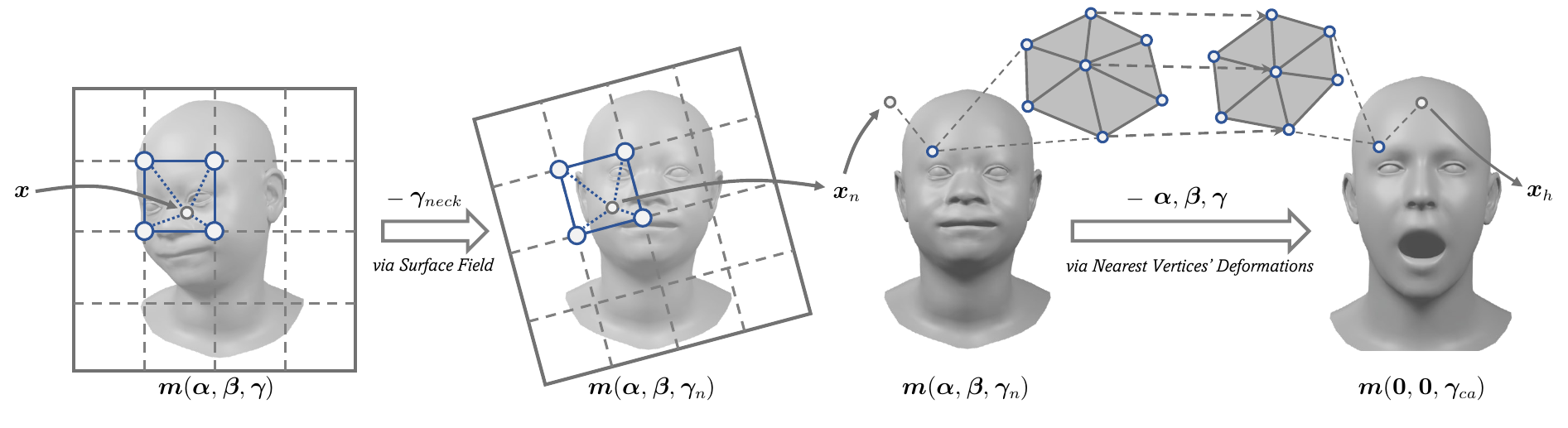}\\
	\captionsetup{type=figure,font=small}
	\vspace{-8pt}
	\caption{Overview of the part-wise 3D deformation field $\mathcal{D}$ in GenHead. We first derive deformation caused by neck pose via the Surface Field approach~\cite{bergman2022generative}. The deformation of an arbitrary 3D point is obtained via tri-linear interpolation between those of pre-defined voxel grids. Then, we eliminate the deformation caused by shape and expression variations, via weighted deformations of its nearest vertices on the FLAME mesh. Here, we only show the derivation of the head region deformation for an illustration.}
	\label{fig:deformation}
\end{strip}

\section{Overview}
We first present more implementation details in Sec.~\ref{sec:implement}, including those of GenHead, 4D data synthesis, and the one-shot 4D head reconstruction pipeline. Then, we provide evaluations of GenHead and additional results of one-shot 4D head synthesis in Sec.~\ref{sec:supp_results}. Finally, we discuss limitations and ethics consideration in Sec.~\ref{sec:discuss}.

\section{More Implementation Details}\label{sec:implement}

\subsection{Part-wise Generative Head Model}\label{sec:genhead_supp}
In this section, we describe more details about the GenHead model, including the shape-aware canonical triplane generator, the part-wise deformation field, the image rendering process, and the learning strategy. We also provide details about data preprocessing and training.

\paragraph{Shape-aware canonical triplane generator.} As described in Sec.~\ref{sec:genhead}, our canonical tri-plane generator $\mathrm{G}_{ca}$ also takes the shape code $\bm \alpha$ as input for synthesizing shape-related canonical appearance. To achieve this, we simply concatenate $\bm \alpha$ with the random noise $\bm z$, and send them together into the mapping network of $\mathrm{G}_{ca}$'s StyleGAN2 backbone. Considering that the shape and appearance only have weak correlations, we randomly replace $\bm \alpha$ sent into the mapping net with an arbitrary shape code at a possibility of $50\%$ during training to avoid overfitting.

\begin{figure}[t]
	\small
	\centering
	\includegraphics[width=1.0\columnwidth]{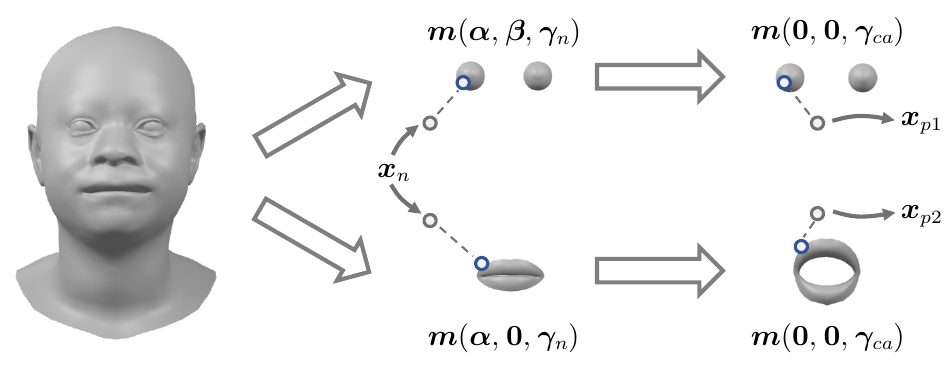}
	\caption{We further derive part-region deformations using the cropped-out eye balls and lip meshes, to better deal with motions of eyes and teeth.}
	\label{fig:deformation_part}
\end{figure}

\paragraph{Part-wise 3D deformation field.} The part-wise 3D deformation field $\mathcal{D}$ produces observation-to-canonical deformations $[\Delta \bm x_{h}, \Delta \bm x_{p}]$ for a 3D point $\bm x$, for modeling shape deformations, as well as animations of neck, face, eyes, and mouth. Illustrations are in Fig.~\ref{fig:deformation} and~\ref{fig:deformation_part}  and we describe the details below.

We first calculate the deformation caused by neck joint rotation $\bm \gamma_{neck}$. We leverage Surface Field (SF) proposed by \cite{bergman2022generative}, which derives the canonical coordinate $\bm x_n$ via
\begin{equation}
    \bm x_n = \bm t_{\bm x}^n\cdot[u,v,w]^\intercal + \langle \bm x - \bm t_{\bm x}\cdot[u,v,w]^\intercal,\bm n_{\bm t_{\bm x}}\rangle \bm n_{\bm t_{\bm x}}^n,
\end{equation}
where $\bm t_{\bm x}$ is $\bm x$'s nearest triangle on the mesh $\bm m(\bm \alpha,\bm\beta,\bm\gamma)$, $[u,v,w]$ is the barycentric coordinates of $\bm x$'s projection onto the triangle, and $\bm n_{\bm t_{\bm x}}$ is the surface normal. $\bm t_{\bm x}^n$ and $\bm n_{\bm t_{\bm x}}^n$ are the corresponding triangle and surface normal on a mesh $\bm m(\bm \alpha,\bm\beta,\bm\gamma_n)$ with canonical neck pose, that is, $\bm\gamma_n=[\bm 0,\bm\gamma_{jaw},\bm\gamma_{eye}]$. 
In practice, we avoid direct SF calculation for every 3D point. Instead, we introduce pre-defined low-resolution voxel grids for SF computation, and 
 approximate the deformation of an arbitrary 3D point with tri-linear interpolation of those of pre-defined voxel grids, as shown in Fig.~\ref{fig:deformation}. This approximation largely reduces the computational cost produced by nearest triangle search. Moreover, the tri-linear interpolation serves as a low-pass filter which largely alleviates the discontinuity of deformations around hairs that have contact with both the face and the shoulder.

After eliminating the neck pose, we tackle the deformation produced by $\bm \alpha$, $\bm \beta$, and $\bm \gamma_{jaw}$ to deform the 3D point to the head canonical space. Specifically, for the 3D point $\bm x_n$ after neck pose canonicalization, we search for its nearest vertex $\bm v_n$ on the mesh $\bm m(\bm \alpha,\bm\beta,\bm\gamma_n)$ and further obtain the one-ring neighborhood $\mathcal{N}(\bm v_n)$ of the vertex. We then calculate the deformation of $\bm x_n$ via weighted summation of the offsets produced by the neighborhood:
\begin{equation}
\Delta (\bm x_n) = \frac{1}{Z}\sum_{\bm v_i \in \mathcal{N}(\bm v_n)}{\omega_i\cdot(\bm v_i^{ca}-\bm v_i)},  \label{eq:deform_face1}
\end{equation} 
where $\bm v_i^{ca}$ denotes $\bm v_i$'s corresponding vertex on mesh $\bm m(\bm 0,\bm 0,\bm \gamma_{ca})$, $\omega_i = 1/\|\bm x_n - \bm v_i\|_2$ is the weighting coefficient 
proportional to the inverse distance between the 3D point and the vertices, and $Z$ is a normalizing scalar. The coordinate $\bm x_h$ in the head canonical space can then be obtained via $\bm x_h = \Delta (\bm x_n) + \bm x_n$. That is, the observation-to-head-canonical deformation $\Delta \bm x_{h} = \Delta (\bm x_n) + \bm x_n - \bm x$.

Finally, we tackle the eye balls' rotation as well as relative movements between lips and teeth. As shown in Fig.~\ref{fig:deformation_part}, we crop out the eye region and lip region from the FLAME mesh. For the eye region, we search for the closest vertices on the eye balls for the 3D point $\bm x_n$, and 
 follow the same procedure as described in Eq.~\eqref{eq:deform_face1} to obtain canonical point $\bm x_{p1}$. 
 For teeth, we notice that their motions are related only to the jaw movements~\cite{garbin2022voltemorph}. Therefore, we derive their deformation via an expressionless lip-region mesh $\bm m(\bm \alpha,\bm 0, \bm \gamma_n)$. We use the offsets between the vertices on this lip mesh and the corresponding vertices on the canonical mesh $\bm m(\bm 0,\bm 0, \bm \gamma_{ca})$ to derive the deformation via Eq.~\eqref{eq:deform_face1}, and obtain the canonical point $\bm x_{p2}$. Therefore, the observation-to-part-canonical deformation $\Delta \bm x_{p} = \bm x_{p1} - \bm x $ or $\bm x_{p2} - \bm x $, where we use $\bm x_{p1} - \bm x$ for points inside the bounding boxes of the eye balls, and $\bm x_{p2} - \bm x$ for the remainings.

\paragraph{Image rendering.} Given the part-wise deformations $[\Delta \bm x_{h}, \Delta \bm x_{p}]$, we can obtain the corresponding features $\bm f_{h}$ and $\bm f_{p}$ from the triplanes $T_{h}$ and $T_{p}$, respectively. An MLP then decodes the features to their radiance $(\sigma_{h},\bm c_{h})$ and $(\sigma_{p},\bm c_{p})$, where $c_{\cdot} \in \mathbb{R}^{32}$ is a color feature and its first three dimensions correspond to $RGB$, as in~\cite{chan2021efficient}. We perform volume rendering~\cite{mildenhall2020nerf,kajiya1984ray} to obtain two feature maps $I_{h}$ and $I_{p}$ via the following equation:
\begin{equation}
% \small
\label{eq:render}
	I({\bm r})  
	=   \sum_{i=1}^{N}t_i(1-{\rm exp}(-\sigma_i\delta_i))\bm c_i, t_i={\rm exp}(-\sum_{j=1}^{i-1}\sigma_j\delta_j),
\end{equation}
where $i$ is the point index along ray $\bm r$ from near to far, and $\delta$ denotes adjacent point distance. We blend the two feature maps to a single foreground feature map $I_{f}$ via the rendered FLAME mask at the same view point:
\begin{equation}
% \small
\label{eq:blend}
	I_f = I_h\odot(1-M_p) + I_p\odot M_p,
\end{equation}
where $\odot$ is element-wise multiplication and $M_p$ is the mask of eyes and inner mouth obtained via rasterization of the FLAME mesh $\bm m(\bm\alpha,\bm\beta,\bm\gamma)$. Similarly, we can obtain the foreground opacity image $I_{opa}$ by setting $\bm c_i$ of all points to $1$ in Eq.~\eqref{eq:render}. Then, we fused the foreground with a 2D background feature map $I_{bg}$ generated by another StyleGAN2:
\begin{equation}
% \small
\label{eq:fuse}
	I_{lr} = I_f\odot I_{opa} + I_{bg}\odot (1-I_{opa}). 
\end{equation}
Finally, the obtained low-resolution feature map $I_{lr}$ is sent into a 2D super-resolution module~\cite{chan2021efficient} to synthesize the final image $I$.

\paragraph{Learning strategy.} We adopt the recent 3D-aware GAN training framework~\cite{chan2021efficient,sun2023next3d} to learn GenHead using monocular real images. During training, we randomly sample $(\bm \alpha, \bm \beta, \bm \gamma)$ and camera pose $\bm \theta$ extracted from the training set, as well as noise code $\bm z$ from normal distribution, and enforce the GenHead model $\mathrm{G}$ to generate a corresponding image $I$. An extra discriminator $\mathrm{D}$ then takes the generated image $I$ as well as a real one $\bar{I}$ from the training set to conduct image-level adversarial learning~\cite{goodfellow2014generative,karras2019style}:
\begin{equation}
\label{eq:adv_2d}
\small
\begin{aligned}
\mathcal{L}_{adv}& = \mathbb{E}_{\bm \alpha,\bm \beta,\bm \gamma,\bm z, \bm \theta}[f(\mathrm{D}(I_{cat}))] \\
& +\mathbb{E}_{\bar{I} \sim p_{real}}[f(-\mathrm{D}(\bar{I}_{cat}))+\lambda\|\nabla \mathrm{D}(\bar{I}_{cat})\|^2],
\end{aligned}
\end{equation}
where $f(u)=\log(1+\exp{(u)})$ is the Softplus function and $|\nabla \mathrm{D}(\cdot))\|^2$ denotes the R1 regularization~\cite{mescheder2018training}.
$I_{cat} = [I,I_{lr},I_{opa},U]$ is a concatenation of the synthesized images $I$, $I_{lr}$, the opacity image $I_{opa}$, and a rasterized correspondence map from $\bm m(\bm\alpha,\bm\beta,\bm\gamma)$ similarly as in~\cite{sun2023next3d}. $\bar{I}_{cat}$ is the corresponding concatenation of $I_{cat}$, where $\bar{I}_{opa}$ is predicted by~\cite{cheng2021per} and $\bar{U}$ is obtained from a reconstructed mesh using~\cite{deng2019accurate,bfmtoflame} and an extra landmark-based optimization step. The additional concatenation of the opacity images and the correspondence maps help with better foreground-background separation and more accurate deformation control.

Besides, we introduce a part-region density regularization to encourage the GenHead to leverage $T_{p}$ instead of $T_{h}$ for generating eyes and inner mouth:
\begin{equation}
    \mathcal{L}_{part} = \sum_{\Pi(\bm x) \in M_p} \sigma_{h}(\bm x),
\end{equation}
where $\Pi(\bm x)$ is the 2D projection of point $\bm x$ in the observation space, and $\sigma_{h}$ is the corresponding volume density obtained from $T_{h}$.

\paragraph{Data preprocessing.} We re-align the FFHQ~\cite{karras2019style} dataset to ensure that the heads have nearly identical scales and are centered in the images. Specifically, we detect facial landmarks of all images using~\cite{yu2021heatmap}. Then, we re-scale and re-center all heads in the images by performing similarity transforms computed between the detected landmarks and the canonical landmarks of BFM~\cite{paysan20093d}. Finally, we center-crop all images and resize them to a resolution of $512^2$. Note that we preserve roll angles of the heads instead of eliminating them as done in~\cite{karras2019style,chan2021efficient}. We use Deep3DRecon~\cite{deng2019accurate} to extract BFM coefficients from the images and transfer them to FLAME codes via~\cite{bfmtoflame}. To improve 3D-to-2D alignment, we conduct extra landmark-based optimization to update the 3D shapes, expressions, eye rotations, and 3D translations. The optimized FLAME codes as well as the camera parameters are used for image synthesis and correspondence map rasterization during training. The camera intrinsics are set identical across all images. In addition, we re-balance the pose distribution of FFHQ based on the estimated head rotations. We duplicate the images by factors of $2$, $4$, $8$, and $16$ for those with yaw angles in ranges of $15^\circ\sim30^\circ$, $30^\circ\sim45^\circ$, $45^\circ\sim60^\circ$, and larger than $60^\circ$, respectively. We also flip all images and extract the corresponding FLAME parameters. This lead to $210$K training images in total compared to the original FFHQ with $70$K images. 

\paragraph{More training details.} 
We randomly sample $\bm \alpha, \bm \beta, \bm \gamma, \bm \theta$ extracted from a same image, and combine them with a random noise $\bm z$. We perform volume rendering at a resolution of $64^2$, and use hirearchical sampling strategy~\cite{mildenhall2020nerf,chan2021efficient} with $48$ coarse sampling points and $48$ fine points. 
We train $\mathrm{G}_{ca}$ and $\mathrm{D}$ via $\mathcal{L}_{adv}$ and $\mathcal{L}_{part}$ to see $25$M images in total. The balancing weights for the two losses are set to $1$ and $10$, respectively. We use Adam optimizer~\cite{kingma2015adam} with $(\beta_1,\beta_2)=(0,0.99)$ and learning rates of $0.0025$ and $0.002$ for the generator and the discriminator, respectively, and set the batch size to $32$. Experiments are conducted on 4 Tesla A100 GPUs with $40$GB memory, and the training takes around $2$ weeks.

\subsection{4D Data Synthesis}\label{sec:data_supp}
In this section, we describe the data synthesis details for training the 4D head reconstruction pipeline. 

Specifically, we follow the data preprocessing procedure described in Sec.~\ref{sec:genhead_supp} to extract $(\bm \alpha, \bm \beta, \bm \gamma, \bm \theta)$ from images in FFHQ and VFHQ. For the dynamic data, we sample $\bm \alpha$ extracted from the FFHQ images and $(\bm \beta, \bm \gamma)$ from the VFHQ images. For the static data, $\bm \alpha, \bm \beta, \bm \gamma$ are sampled from both the FFHQ and VFHQ images. Note that for $\gamma_{neck}$, we sample it from a manually-defined distribution, with pitch in $[-0.2,0.2]$ rad, yaw in $[-0.5,0.5]$ rad, and roll in $[-0.1,0.1 ]$ rad. For the camera pose $\bm \theta$, we also sample it from a pre-defined uniform distribution that covers most of the camera parameters estimated from FFHQ, with pitch in $[-0.25,0.65]$ rad, yaw in $[-0.78,0.78]$ rad, and roll in $[-0.25,0.25]$ rad. The camera radius are uniformly sampled from $[3.65,4.45]$, and the camera look-at position from $[-0.01,0.01]\times[-0.01,0.01]\times[0.02,0.04]$. We use fixed camera intrinsics similarly as in GenHead, with a field of view (FoV) equal to $12^\circ$. 

For a certain identity $(\bm \alpha, \bm z)$ with different motions and camera poses, we use the same $\bm z$ to generate the background image. We maintain the intermediate outputs as additional supervisions as described in Sec.~\ref{sec:data}. For the sampled triplane features $\bar{T}(\bm x)$, we randomly choose $4000$ coarse sampling points during the rendering process and record their features $\bm f_{h}$ from the triplanes $T_{h}$. Besides, for all synthesized images $\bar{I}_{re}$, we use an average $\mathcal{W}$ space~\cite{abdal2019image2stylegan} vector for the modulated convolutional layers in the 2D super-resolution module. Visualizations of the synthetic data can be found in Sec.~\ref{sec:data_vis}.

\subsection{Animatable Triplane Reconstructor}

\begin{figure*}[p]
    \small
	\centering
	\includegraphics[width=1.0\textwidth]{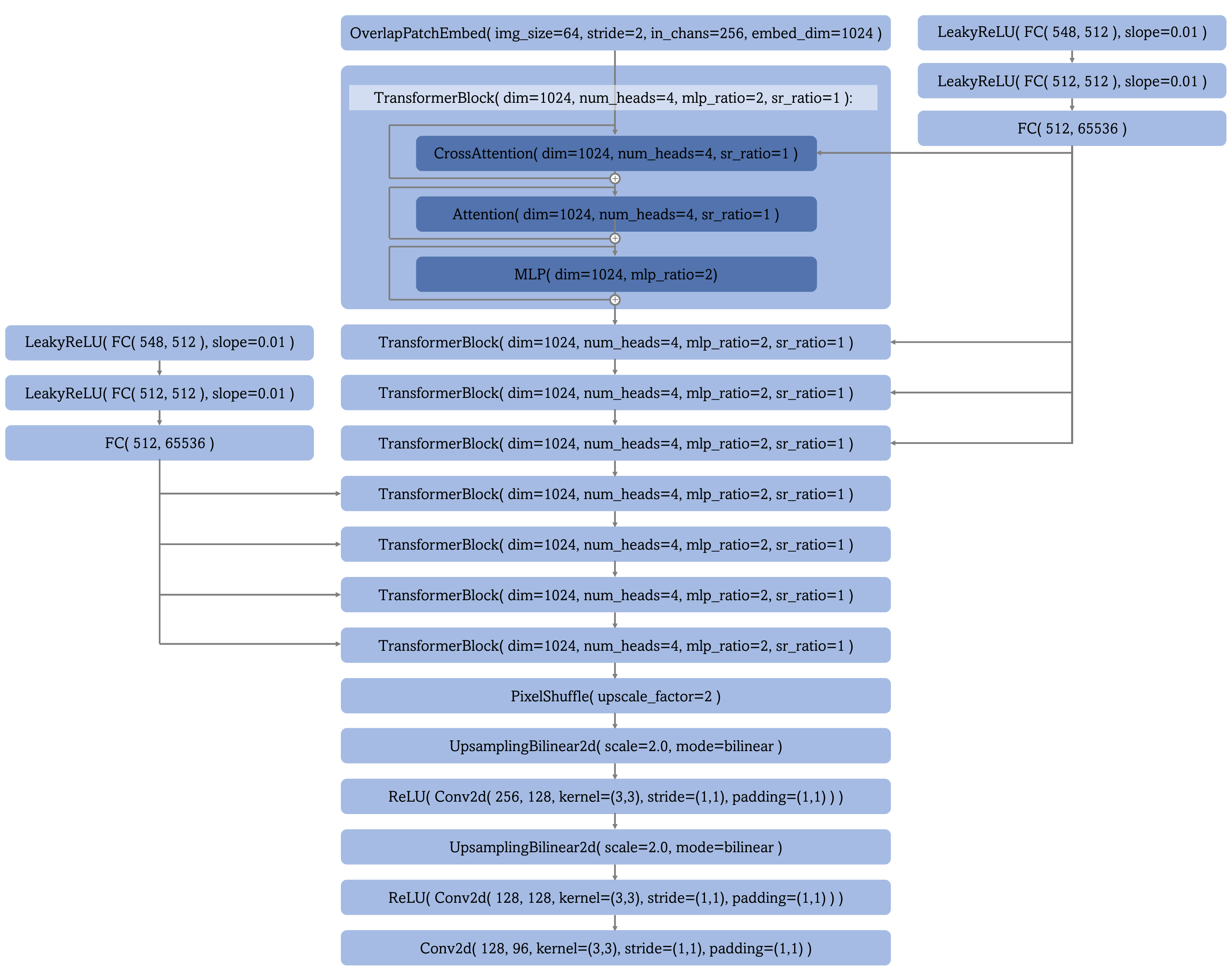}
	
	\caption{Network structure of the canonicalization and reenactment module $\Phi$.\label{fig:phi}}
	
% 	\vspace{-8pt}
\end{figure*}

\paragraph{Canonicalization and reenactment module.} The canonicalization and reenactment module $\Phi$ consists of a de-expression module $\Phi_{de}$ and a reenactment module $\Phi_{re}$ sharing the same structure. They each has four transformer blocks with a cross-attention layer, a self-attention layer, and an MLP. An extra MLP is utilized to expand the spatial dimension of the motion feature $\bm v$ for computing the cross-attentions. A detailed structure of $\Phi$ can be found in Fig.~\ref{fig:phi}.

\begin{figure*}[t]
	\small
	\centering
	\includegraphics[width=0.95\textwidth]{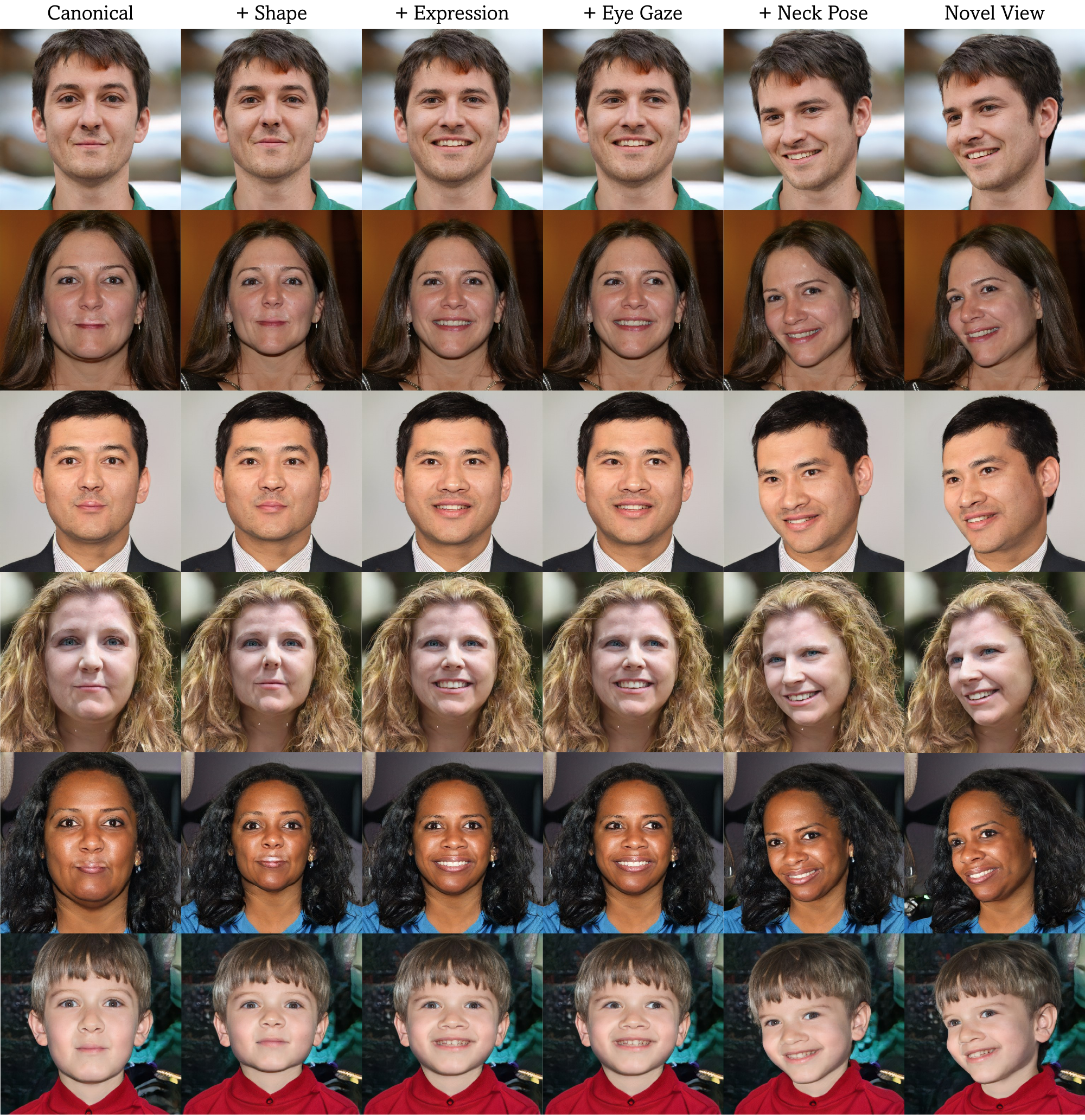}
 \vspace{-3pt}
	\caption{Head images synthesized by GenHead. We can generate diverse virtual identities and support individual control over head shapes, expressions, eye gazes, neck poses, and camera poses. \textbf{Best viewed with zoom-in.}}
	\label{fig:genhead_main}
 \vspace{-3pt}
\end{figure*}

\paragraph{Motion feature.} We utilize the motion features from~\cite{wang2023progressive} as input to $\Phi$. More specifically, we use a concatenation of the features from~\cite{wang2023progressive}'s expression encoder $\mathrm{E}_{exp}$, lip encoder $\mathrm{E}_{lip}$, and eye encoder $\mathrm{E}_{eye}$. This leads to a motion vector $\bm v$ of dimension $30+512+6=548$.

\begin{figure*}[p]
    \small
	\centering
	\includegraphics[width=1.0\textwidth]{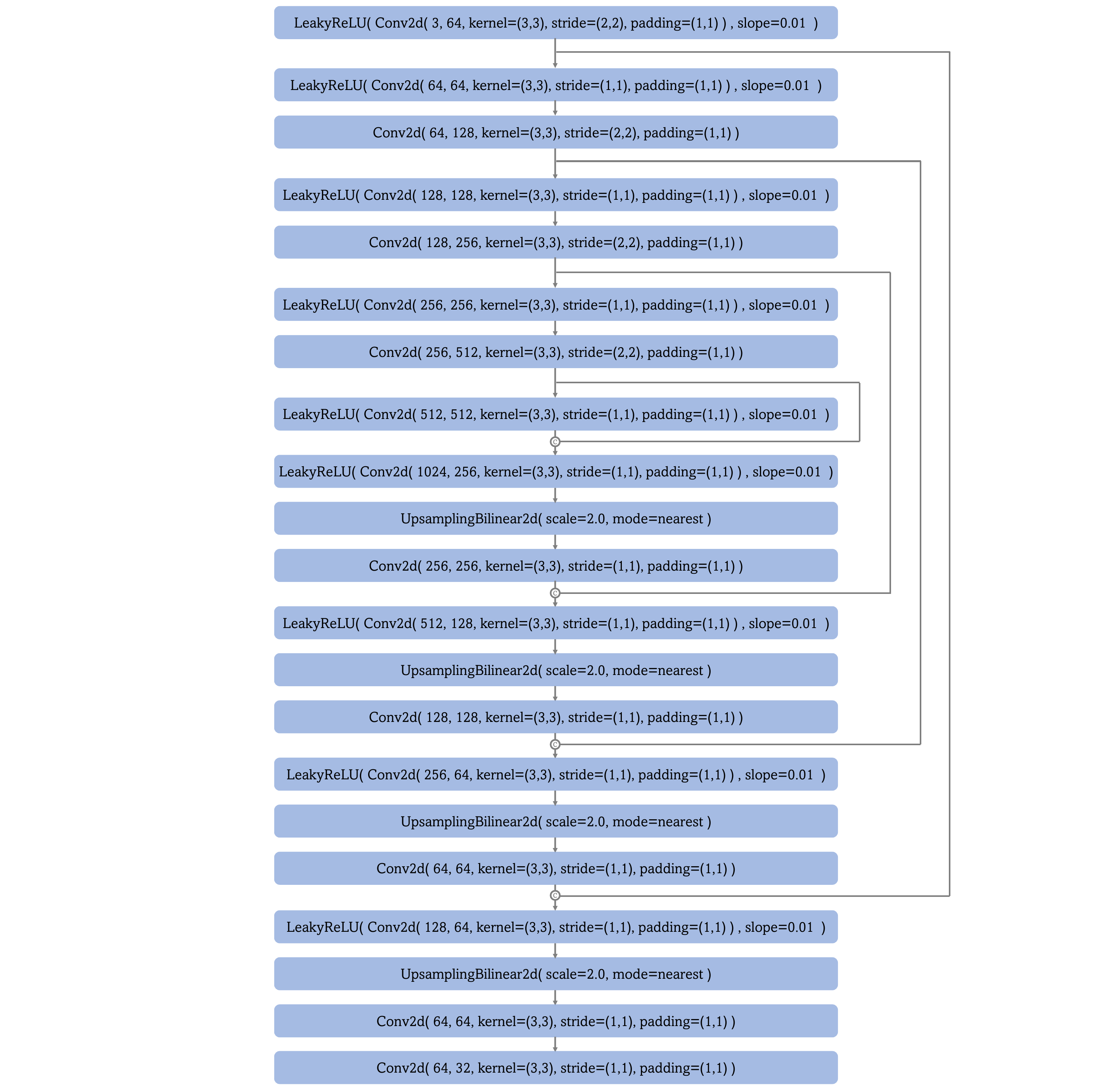}
	
	\caption{Structure of the background U-Net.\label{fig:bg}}
	
% 	\vspace{-8pt}
\end{figure*}

\paragraph{Background prediction.} We leverage a U-Net to predict the background feature map $I_{bg}$ from the input image. The structure of it is illustrated in Fig.~\ref{fig:bg}. 

\paragraph{Image rendering.} We follow a similar procedure as in Sec.~\ref{sec:genhead_supp} for image synthesis. Specifically, a foreground feature map $I_{f}$ is rendered from the reconstructed triplane $T$ via Eq.~\eqref{eq:render}, and fused with the predicted background $I_{bg}$ from the U-Net via Eq.~\eqref{eq:fuse}. Note that for the 4D head synthesis pipeline, we do not use part-wise triplanes as in GenHead but a single triplane $T$ to represent the whole head region and render the corresponding images.

\paragraph{More training details.} We train the animatable triplane reconstructor $\Psi$ using the synthetic data described in Sec.~\ref{sec:data_supp} and Sec.~\ref{sec:data}, as well as the training objective in Sec.~\ref{sec:training}. We initialize the projection weights of all cross-attention layers to zeros, and use the pre-trained weights from GenHead for the radiance decoding MLP and 2D super-resolution module in the renderer $\mathcal{R}$, as well as the discriminator $\mathrm{D}$. The balancing weights for each loss term in Eq.~\eqref{eq:training} are set to $1$, $1$, $0.1$, $1$, $0.3$, $1$, and $0.01$ for $\mathcal{L}_{re}$, $\mathcal{L}_{f}$, $\mathcal{L}_{tri}$, $\mathcal{L}_{depth}$, $\mathcal{L}_{opa}$, $\mathcal{L}_{id}$, and $\mathcal{L}_{adv}$, respectively. During the first $1000$K images, we do not use $\mathcal{L}_{adv}$ and fix the network parameters inside the renderer $\mathcal{R}$. After seeing $1000$K images, we activate $\mathcal{L}_{adv}$ and unfrozen the trainable parameters in $\mathcal{R}$. We discard $\mathcal{L}_{f}$, $\mathcal{L}_{tri}$, and $\mathcal{L}_{depth}$ at this stage. Following GenHead, we perform volume rendering with $48$ coarse sampling points and $48$ fine points per ray. We use a volume rendering resolution of $64^2$ at the first $1000$K images, and gradually increase the resolution to $128^2$ at the next $1000$K images. During the entire training process, we use a fixed average $\mathcal{W}$ space vector from GenHead for the 2D super-resolution module as in Sec.~\ref{sec:data_supp}. 

We train $\Psi$ and $\mathcal{R}$ to see $12$M images in total. We use Adam optimizer with $(\beta_1,\beta_2)=(0.9,0.999)$ and a learning rate of $1e-4$ for all the networks. The batch size is set to $32$, half of which are dynamic data and half of which are static data. The model is trained with 8 Tesla A100 GPUs with $80$GB memory for $10$ days.

\section{More Results}\label{sec:supp_results}
\subsection{Evaluation of GenHead}
\label{sec:supp_results_genhead}

\paragraph{Controllable head image generation.} Figure~\ref{fig:genhead_main} shows the controllable head image generation results of GenHead. We start from canonical appearances synthesized by random noise $\bm z$ with an average shape and neutral expression (\ie, $\bm \alpha, 
\bm \beta, \bm \gamma =\bm 0$). Then, we introduce shape variations as well as expression and pose control to different canonical heads. As shown, GenHead supports individual control over head shape, expression, eye gaze, neck pose, and camera pose. The synthesized images are of high photorealism and can be directly used as training data to facilitate our one-shot 4D head reconstruction pipeline.

\paragraph{Shape-aware canonical appearance.}

\begin{figure}[t]
	\small
	\centering
	\includegraphics[width=1.0\columnwidth]{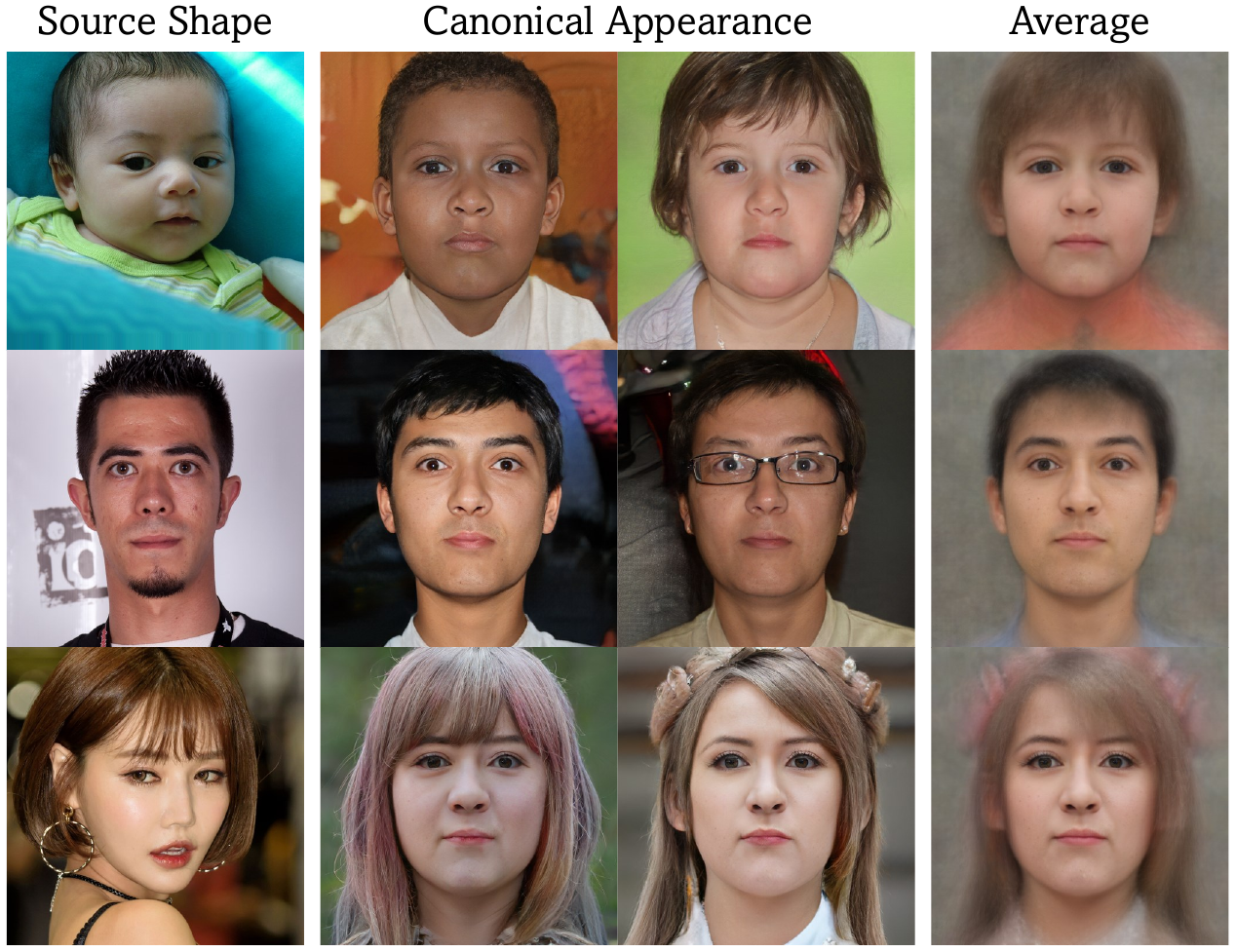}
	\caption{Generated canonical appearances with shape codes from different source images.}
	\label{fig:genhead_app}
 \vspace{-5pt}
\end{figure}

Figure~\ref{fig:genhead_app} shows the synthesized canonical appearances given shape codes $\bm \alpha$ extracted from different source images as condition. In each row, we fix the shape code and vary the random noise $\bm z$, and compute an image-space average appearance as well. 
As shown, the distribution of the canonical appearance is influenced by the given shape code. In the ablation study, we show that this strategy largely improves the image generation quality in terms of FID without sacrificing the controllability. Note that at inference time, we can use different shape codes for the appearance and the deformation field for more diverse virtual head image generation.

\begin{table*}[t]\small
\centering
\caption{Comparisons between head GANs on controllable items, generation quality\&diversity (Q. \& Div.), and control accuracy.}
\vspace{-5pt}
\begin{tabular}{l|ccccc|l|lllll}
\toprule

% \cline{2-9}
\multicolumn{1}{c}{\multirow{2}{*}{Method}} & \multicolumn{5}{c}{Independent Control Item} & \multicolumn{1}{c}{Q. \& Div. }  & \multicolumn{5}{c}{Control Accuracy} \\ 
\cline{2-12} 
 \multicolumn{1}{c}{} & Exp. & Neck& Gaze & Teeth & BG & FID $\downarrow$ &  APD $\downarrow$ & AED $\downarrow$   & LMD $\downarrow$   & ASD $\downarrow$   & ASV $\downarrow$ \\
 \midrule
 DiscoFaceGAN~\cite{deng2020disentangled} &\checkmark&&&&& 12.9 & 0.031 &  0.829 & -  & -  & 66.6  \\ 
3DFaceShop~\cite{tang2022explicitly} &\checkmark&&&&\checkmark& 21.7 & 0.022\tikzcircle[gold,fill=gold]{2pt} & 0.865 & - & -  & 7.3 \\
AniFaceGAN~\cite{wu2022anifacegan} &\checkmark&&&&& 20.1 & 0.039 & 0.687\tikzcircle[gold,fill=gold]{2pt} & -  & -  &  19.9  \\
GNARF~\cite{bergman2022generative} &\checkmark&&&&& 6.6 & - & - &  - & - & - \\ 
OmniAvatar~\cite{xu2023omniavatar} &\checkmark&\checkmark&&&\checkmark& 5.8 & - & - &  - & - & - \\ 
Next3D~\cite{sun2023next3d} &\checkmark&&\checkmark&&& 3.9\tikzcircle[gold,fill=gold]{2pt} & 0.029\tikzcircle[bronze,fill=bronze]{2pt} & 0.868  &  23.1 & 20.8 & 18.6 \\ 
\midrule
A  Baseline EG3D~\cite{chan2021efficient} &&&&&& 4.8 & N/A & N/A & N/A & N/A & N/A \\
B  + Deformation \& BG &\checkmark&\checkmark&&&\checkmark& 6.5 & 0.031 & 0.783  &  9.6 & 9.1 & 9.3 \\
C  + Correspondence Map &\checkmark&\checkmark&&&\checkmark& 7.8  & 0.030 & 0.757 & 10.5 & 9.9 & 5.5\tikzcircle[gold,fill=gold]{2pt} \\
D  + Opacity Image &\checkmark&\checkmark&&&\checkmark& 9.5 & 0.030 & 0.722 & 9.1\tikzcircle[silver,fill=silver]{2pt} & 8.7\tikzcircle[silver,fill=silver]{2pt} & 8.1 \\ % whats's the relationship of eg3d and our baseline?
E  + Shape Condition &\checkmark&\checkmark&&&\checkmark& 4.7\tikzcircle[bronze,fill=bronze]{2pt}  & 0.032 & 0.700\tikzcircle[bronze,fill=bronze]{2pt} & 9.5\tikzcircle[bronze,fill=bronze]{2pt} & 9.0\tikzcircle[bronze,fill=bronze]{2pt} & 6.4\tikzcircle[bronze,fill=bronze]{2pt}   \\
F  + Part Model (Ours) &\checkmark&\checkmark&\checkmark&\checkmark&\checkmark& 4.6\tikzcircle[silver,fill=silver]{2pt}  & 0.028\tikzcircle[silver,fill=silver]{2pt} & 0.699\tikzcircle[silver,fill=silver]{2pt} &  9.1\tikzcircle[gold,fill=gold]{2pt} &  8.5\tikzcircle[gold,fill=gold]{2pt} &  6.1\tikzcircle[silver,fill=silver]{2pt} \\
\bottomrule
\end{tabular}

\label{tab:genhead_main}
\vspace{-8pt}
\end{table*}

\begin{table}[t]\small
\centering
\caption{Comparison on 3D consistency of different head GANs using the evaluation metrics of~\cite{xiang2022gram}.}
\vspace{-5pt}
\begin{tabular}{l|ll}
\toprule

Method  &  PSNR $\uparrow$ &  SSIM $\uparrow$  \\
\midrule
EG3D~\cite{chan2021efficient} & 34.0 & 0.928 \\
Next3D~\cite{sun2023next3d} & 34.5 & 0.941\\
\hline
Ours & 34.2 & 0.940 \\

\bottomrule
\end{tabular}

\label{tab:genhead_consistency}
\vspace{-5pt}
\end{table}

\paragraph{Comparisons with previous head GANs.} We compare GenHead with existing 3D head GANs: DiscoFaceGAN~\cite{deng2020disentangled}, AniFaceGAN~\cite{wu2022anifacegan}, 3DFaceShop~\cite{tang2022explicitly}, GNARF~\cite{bergman2022generative}, OmniAvatar~\cite{xu2023omniavatar}, Next3D~\cite{sun2023next3d}, and EG3D~\cite{chan2021efficient}. Since GNARF and OmniAvatar do not release their codes and models for head generation, we only compare with their reported FID. For a fair comparison, we re-train GenHead using FFHQ images aligned by~\cite{chan2021efficient} instead of our new alignment described in Sec.~\ref{sec:genhead_supp}. We train the model at a resolution of $256^2$ for efficiency. The following ablation study also adopts the same configuration.

Table~\ref{tab:genhead_main} shows the quantitative results. For image synthesis quality, we measure the FID score between 50K generated images and all available real images in the training set. For control accuracy, we measure APD, AED, Landmark Distance (LMD), Average Shape Distance (ASD), and Average Shape Variance (ASV). For APD, we compute the distance between input camera angles and those reconstructed by~\cite{deng2019accurate} using $1000$ generated images. For AED, we manually extract expressions from $30$ reference images with typical and distinct expressions, and combine them with $50$ generated appearances for image synthesis. We leverage EMOCA~\cite{danvevcek2022emoca} to compute AED between the reference images and the synthesized ones. For ASD and LMD, we measure the vertex and landmark distances between input shapes and those reconstructed from synthesized images using EMOCA, respectively. We randomly sample $500$ shape codes from the training set, and generate $10$ different appearances for each shape. Since DiscoFaceGAN, 3DFaceShop, and AniFaceGAN require BFM shapes as condition which differ from our FLAME topology, we only compare with Next3D for these two metrics. For ASV, we calculate the vertex variance between reconstructed shapes of different images synthesized with the same shape code. Similarly, we sample $500$ shape codes and $10$ different appearances for each shape. 

\begin{figure}[t]
	\small
	\centering
	\includegraphics[width=1.0\columnwidth]{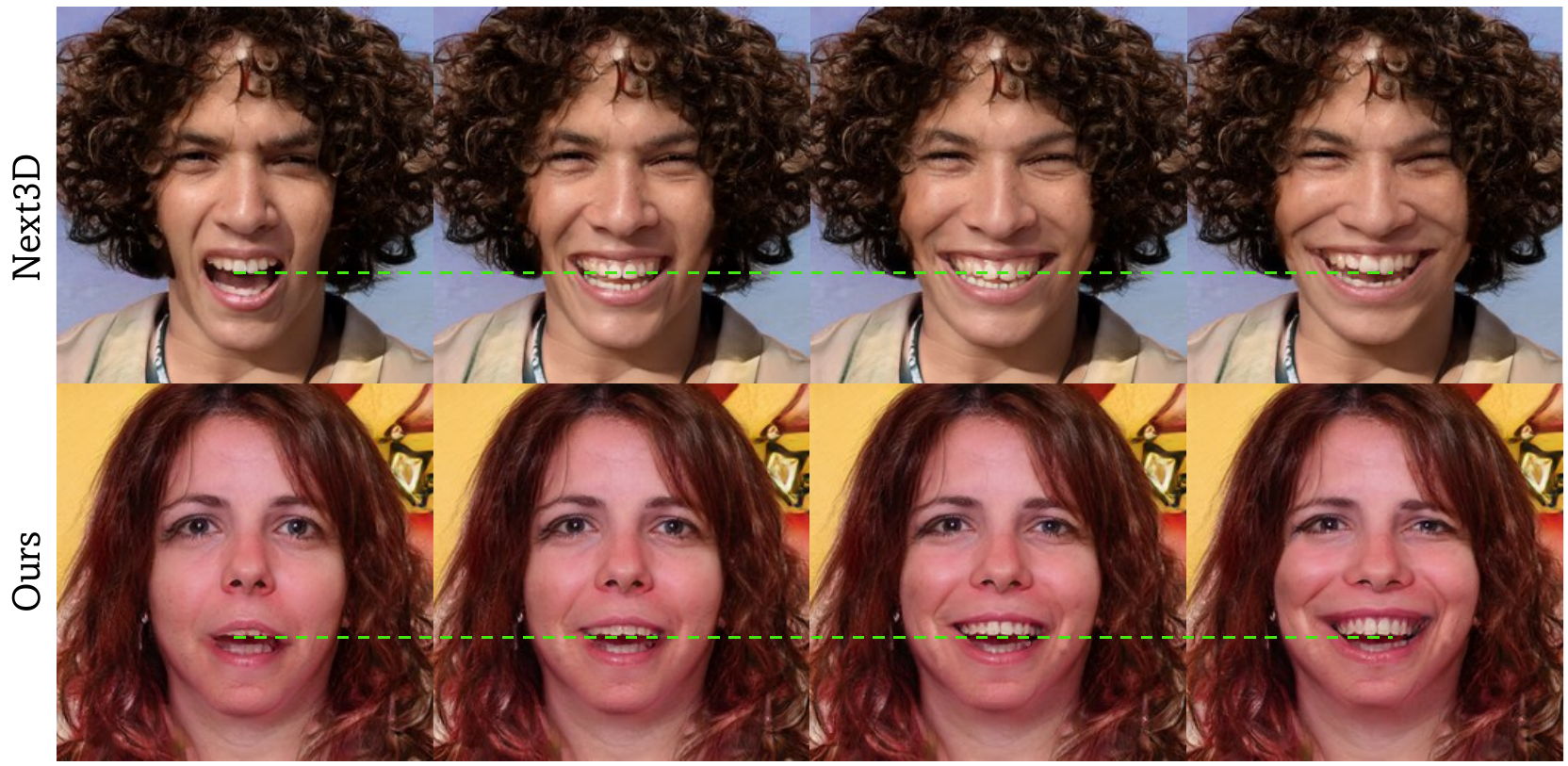}
	\caption{GenHead better captures the relative movements between lips and teeth, where the upper teeth should stay steady under expression changes.}
	\label{fig:teeth}
 \vspace{-5pt}
\end{figure}

As shown, GenHead achieves the best overall control accuracy, with competitive image generation quality. What's more, our method supports full control over expression, neck pose, eye gaze, relative motions between lips and teeth, and background separation, which cannot be achieved by previous methods. Figure~\ref{fig:teeth} shows a comparison between GenHead and Next3D for separate teeth control. Ideally, lip motions should not influence the position of the upper teeth. However, the upper teeth synthesized by Next3D move with the lip variations. By contrast, our method maintains the position of the upper teeth during expression changes, which is more consistent with reality.

In Tab.~\ref{tab:genhead_consistency}, we further compare the 3D consistency of our method with EG3D and Next3D. We use the evaluation metric from GRAM-HD~\cite{xiang2022gram} which measures the reconstruction fidelity of a 3D reconstruction method NeuS~\cite{wang2021neus} on multi-view images generated by different generators. As shown, our method yields comparable results with the two baselines. The high 3D consistency of GenHead guarantees reasonable synthetic 4D data for learning the subsequent one-shot 4D head synthesizer. For further improvement of the 3D consistency, a possible way is to use the 3D-to-2D imitative strategy proposed by Mimic3D~\cite{chen2023mimic3d} to generate high-resolution tri-planes for direct volume rendering. This way, the synthetic 4D data of GenHead will have even higher 3D consistency which can further facilitate the learning of the 4D head synthesizer.

\paragraph{Ablation study.} We conduct ablation studies to validate different components in GenHead. \textit{\textbf{A}} is a baseline identical to EG3D. \textit{\textbf{B}} adds the deformation field (non-part-wise) and the background network on top of \textit{A}. \textit{\textbf{C}} introduces the correspondence map condition to the dual discriminator and \textit{\textbf{D}} further introduces the opacity image condition. \textit{\textbf{E}} leverages the shape condition for synthesizing canonical appearance. \textit{\textbf{F}} utilizes the part-wise deformation field and canonical tri-planes which is our final configuration.

\begin{figure}[t]
	\small
	\centering
	\includegraphics[width=1.0\columnwidth]{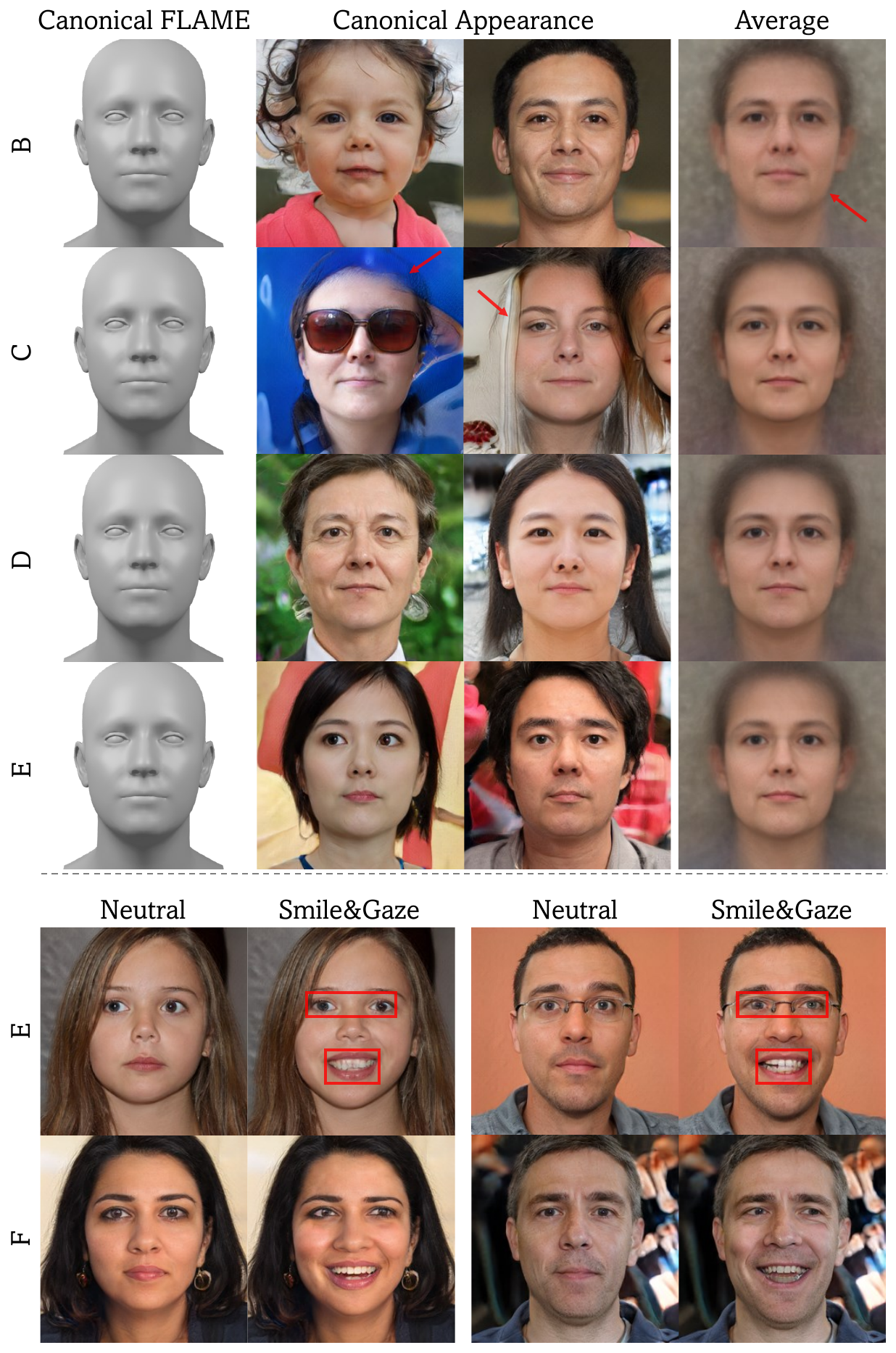}
  \vspace{-15pt}
	\caption{Ablation study. \textbf{Top:} synthesized canonical appearances and their image-space averages. \textbf{Bottom:} synthesized images under expression and gaze variations.}
	\label{fig:genhead_ablation}
 \vspace{-15pt}
\end{figure}

Table~\ref{tab:genhead_main} and Fig.~\ref{fig:genhead_ablation} show the comparisons between different configurations. Naively introducing the 3D deformation (\textit{B}) cannot achieve satisfactory motion control accuracy as indicated by the relatively higher AED, LMD, ASD and ASV, as well as the blurry image-space average of the canonical appearance in Fig.~\ref{fig:genhead_ablation}.
Adding the correspondence map condition (\textit{C}) improves the alignment of the canonical appearances and leads to better controllability in terms of AED and ASV, but sacrifices image quality and leads to transparent foregrounds. These artifacts also lead to inaccurate shape reconstruction results as indicated by the higher LMD and ASD. Further introducing the opacity image (\textit{D}) resolves the transparency issue and improves the overall controllability, but leads to a further quality drop in terms of FID. Conditioning the canonical appearance on shape code (\textit{E}) largely improves the image quality and maintains competitive control accuracy. Finally, adopting the part-wise deformation and canonical tri-planes (\textit{F}) improves the controllability of eye gaze and the quality of inner mouth, and yields the best overall control accuracy.

\subsection{Synthetic 4D Data}\label{sec:data_vis}
\begin{figure*}[p]
    \small
	\centering
	\includegraphics[width=1.0\textwidth]{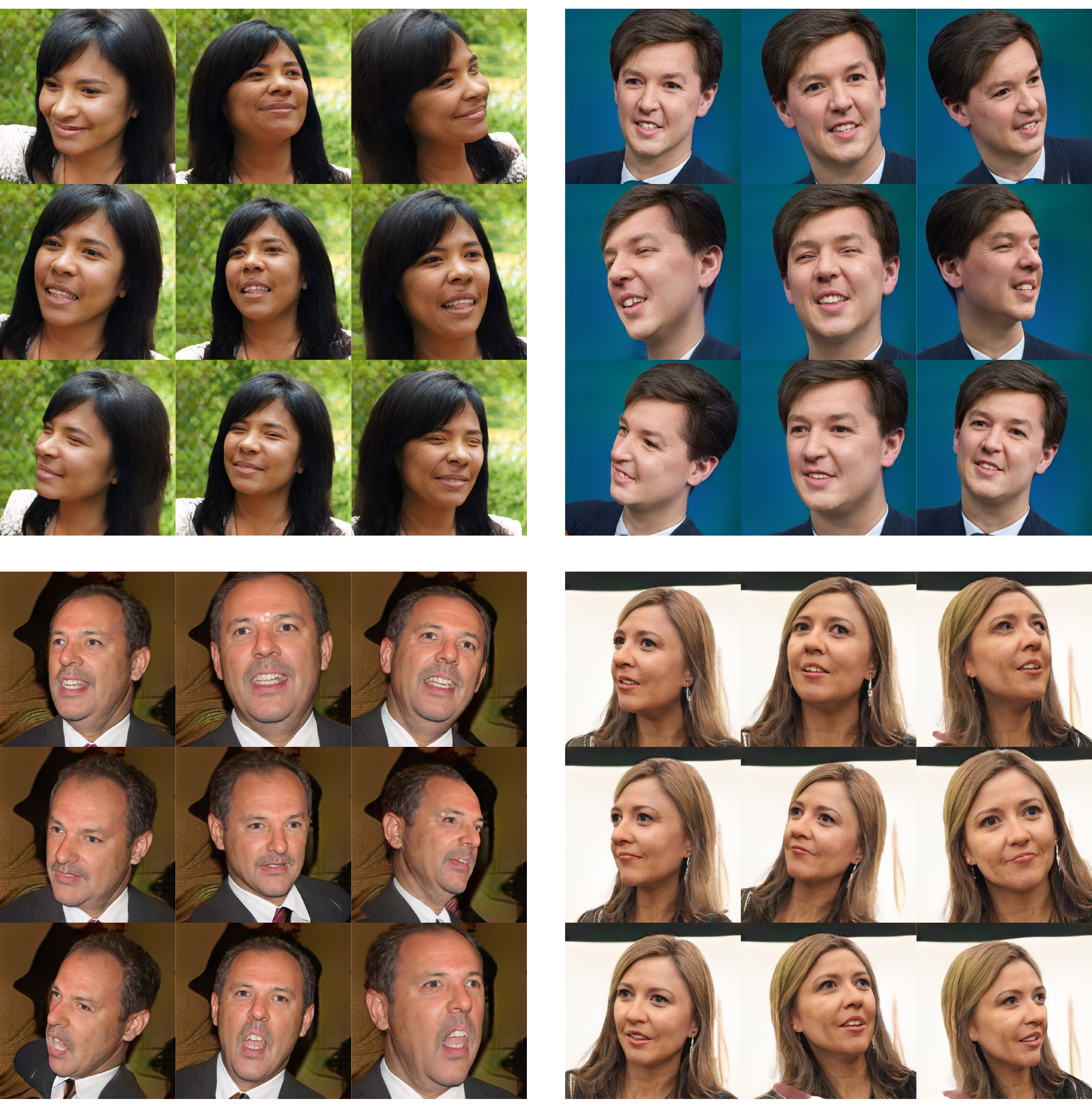}
	
	\caption{Synthesized dynamic data from GenHead for learning one-shot 4D head synthesis.\label{fig:dynamic_data}}

% 	\vspace{-8pt}
\end{figure*}

\begin{figure*}[p]
    \small
	\centering
	\includegraphics[width=1.0\textwidth]{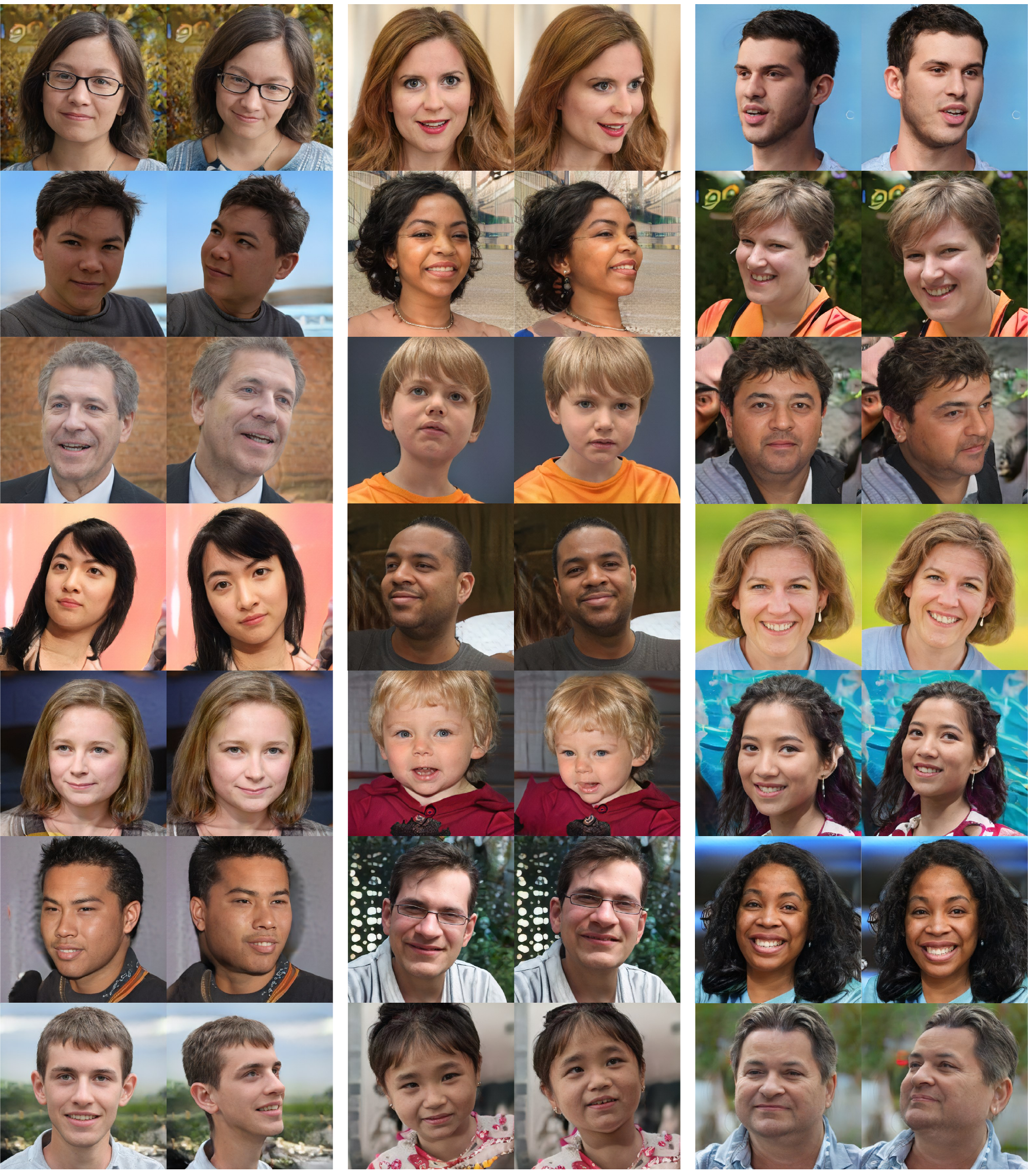}
	
	\caption{Synthesized static data from GenHead for learning one-shot 4D head synthesis.\label{fig:static_data}}

% 	\vspace{-8pt}
\end{figure*}

We showcase our synthetic data for training the one-shot 4D head synthesis pipeline in Fig.~\ref{fig:dynamic_data} and~\ref{fig:static_data}. As shown, the dynamic data contain virtual identities each with different motions and camera poses, while the static data contain pose variations only. The static data have a wider range of identity distribution to enhance the model's generalizability. Backgrounds are fixed for each identity to facilitate learning the foreground-background separation.

\subsection{One-Shot 4D Head Synthesis}
\label{sec:results_main_supp}
\begin{figure*}[p]
    \small
	\centering
	\includegraphics[width=1.0\textwidth]{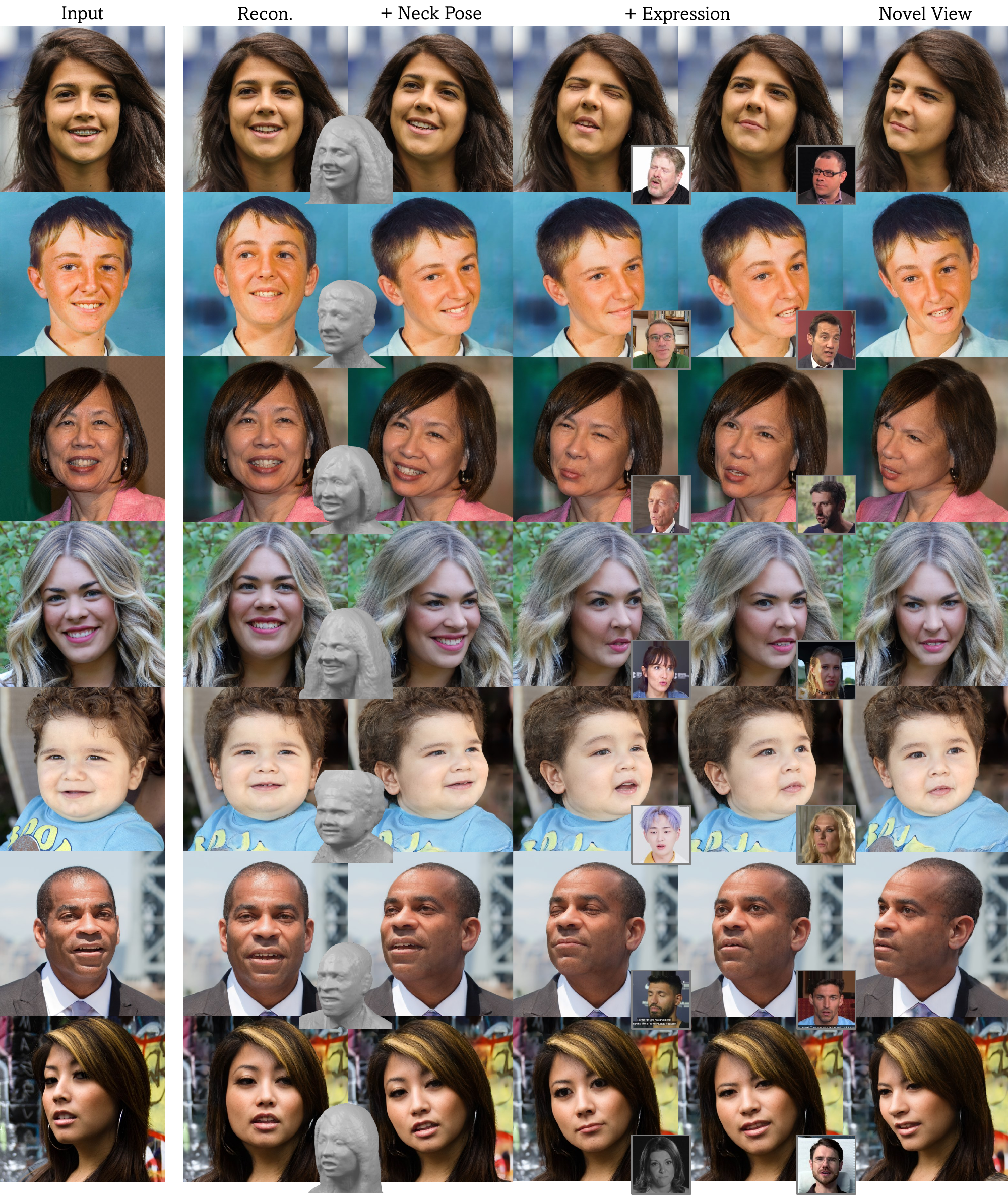}
	
	\caption{One-shot 4D head synthesis results by our method.\label{fig:results_main_supp2}}

% 	\vspace{-8pt}
\end{figure*}

We provide additional one-shot 4D head synthesis results in Fig.~\ref{fig:results_main_supp1} and~\ref{fig:results_main_supp2}. Our method can faithfully reconstruct head avatars from the given portraits and control their expressions and poses for photorealisitic image synthesis.

\subsection{Comparisons with the Prior Art}
\label{sec:compare_supp}
\begin{figure*}[p]
    \small
	\centering
	\includegraphics[width=1.0\textwidth]{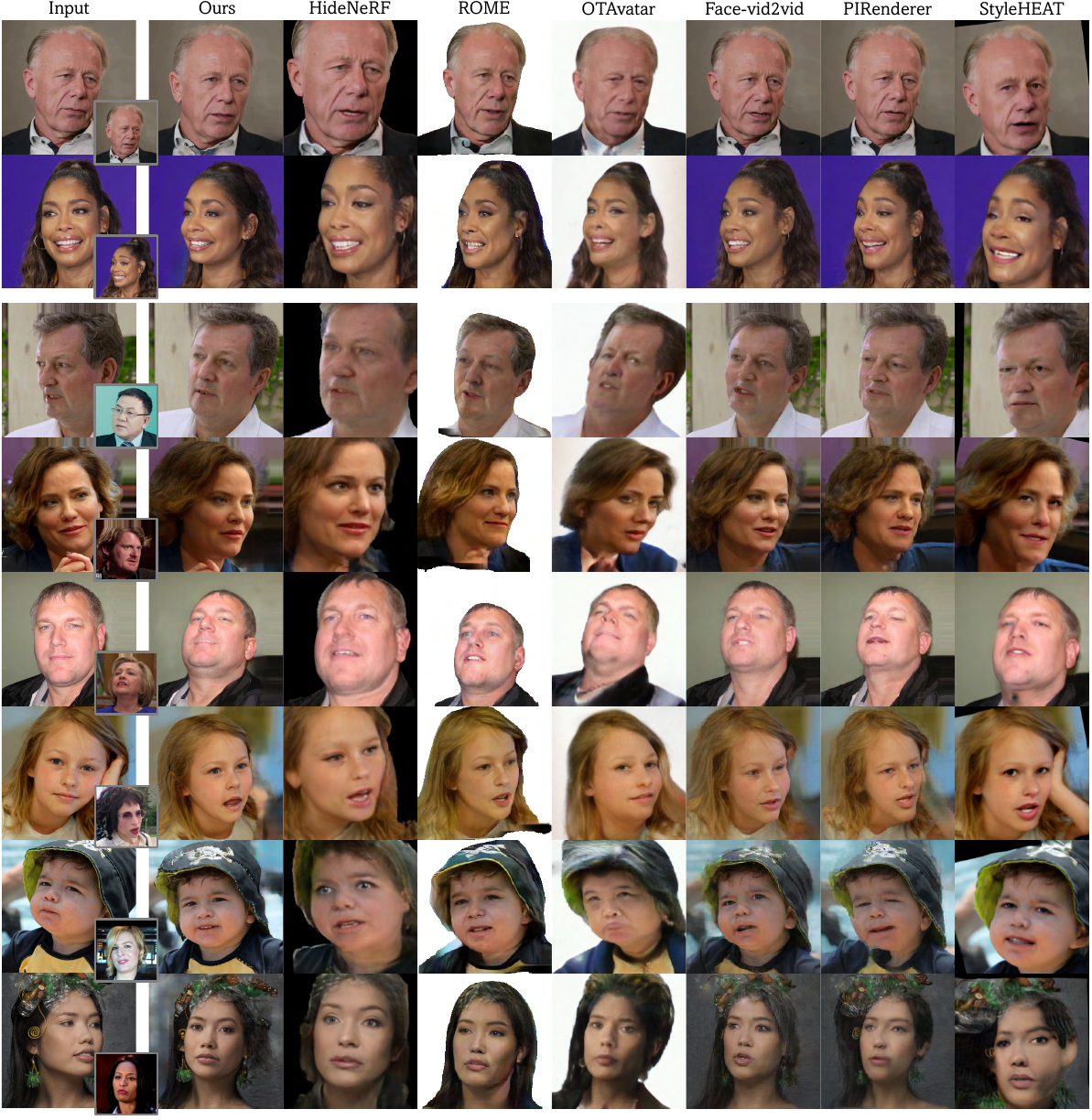}
	
	\caption{Comparison on one-shot head reenactment with previous methods. \textbf{Best viewed with zoom-in.}\label{fig:compare_supp1}}

\end{figure*}

\begin{figure*}[p]
    \small
	\centering
	\includegraphics[width=1.0\textwidth]{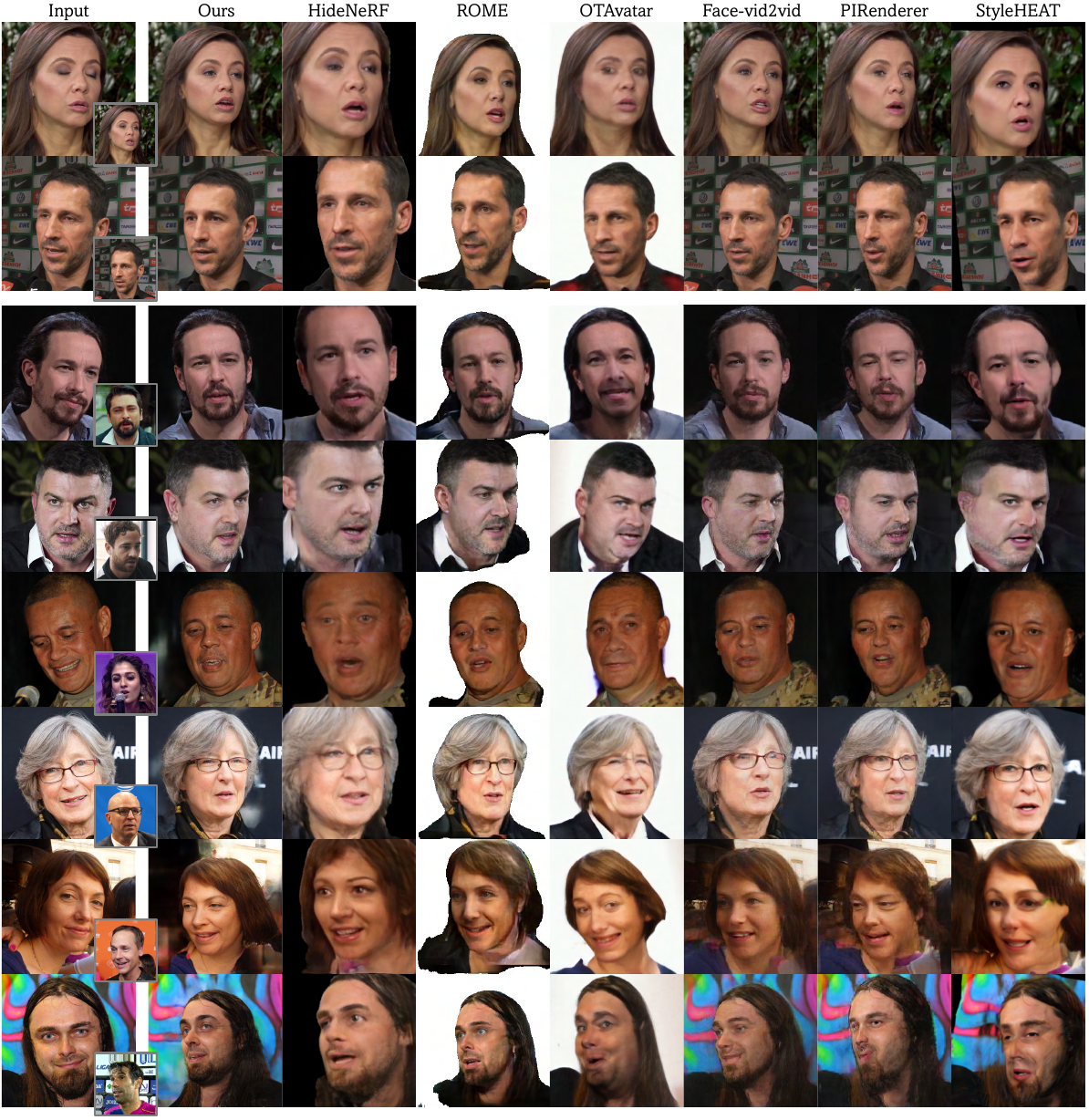}
	
	\caption{Comparison on one-shot head reenactment with previous methods. \textbf{Best viewed with zoom-in.}\label{fig:compare_supp2}}

\end{figure*}

Figure~\ref{fig:compare_supp1} and~\ref{fig:compare_supp2} shows more visual comparisons on one-shot head reenactment with previous methods. Our method yields the best visual quality, and can well preserve the identities and geometries of the source images under large pose variations compared to the alternatives.

\subsection{$\Phi_{de}$ in Different Configurations}

\begin{figure}[t]
	\small
	\centering
	\includegraphics[width=1.0\columnwidth]{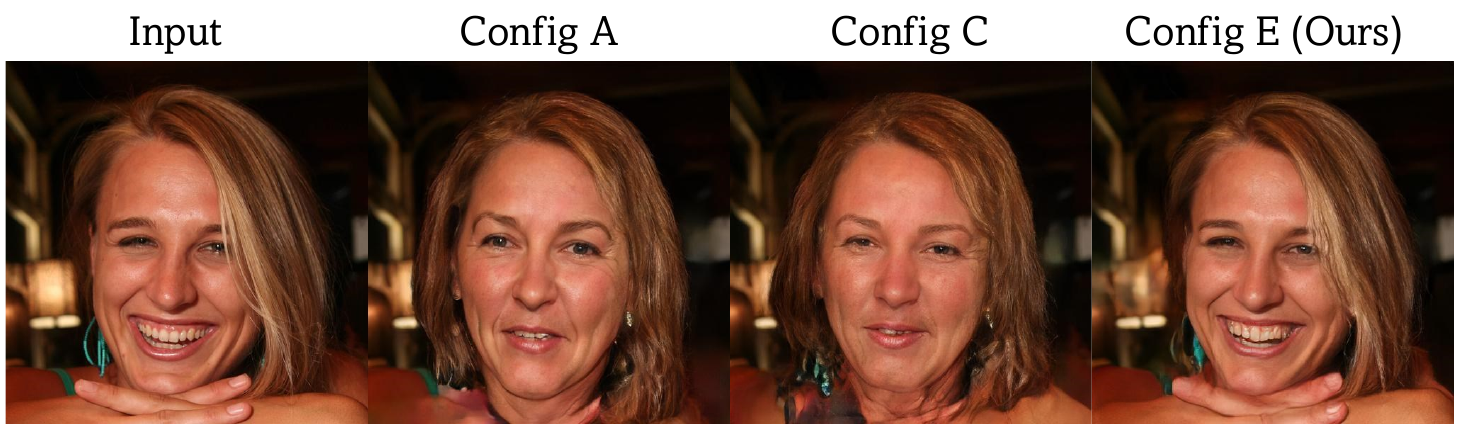}
	\caption{Reconstruction results by switching off all cross-attention layers in $\Phi_{de}$ and $\Phi_{re}$.}
	\label{fig:deexp}

\end{figure}

Figure~\ref{fig:deexp} shows the reconstruction results of an input image by switching off all cross-attention layers in $\Phi_{de}$ and $\Phi_{re}$ in different configurations (corresponding to the ablations in Sec.~\ref{sec:ablation}). Without cross-attentions, config. \textbf{\textit{A}} and \textbf{\textit{C}} still canonicalize expression of the input, which indicates that the self-attention and feed-forward layers in $\Phi_{de}$ take the responsibility of expression neutralization. By contrast, ours handles expression neutralization only through the cross-attentions thus the reconstructed expression is unchanged under this circumstance.

\subsection{Out-of-Distribution Results}

\begin{figure}[t]
	\small
	\centering
	\includegraphics[width=0.8\columnwidth]{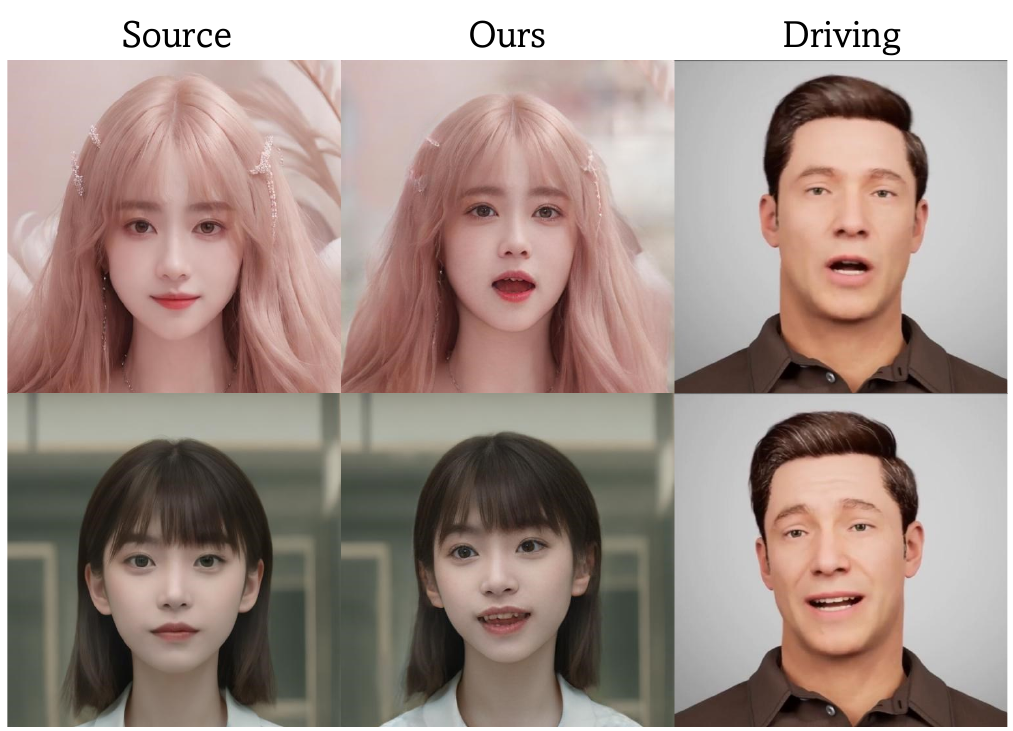}
	\caption{Reenactment results on out-of-distribution subjects.}
	\label{fig:ood}

\end{figure}

We show reenactment results on out-of-distribution subjects in Fig.~\ref{fig:ood}, where the sources are generated from Stable Diffusion~\cite{rombach2022high} and the drivings from Unity Engine. Our method produces reasonable
results on these cases.

\section{Discussions} \label{sec:discuss}

\begin{figure}[t]
	\small
	\centering
	\includegraphics[width=1.0\columnwidth]{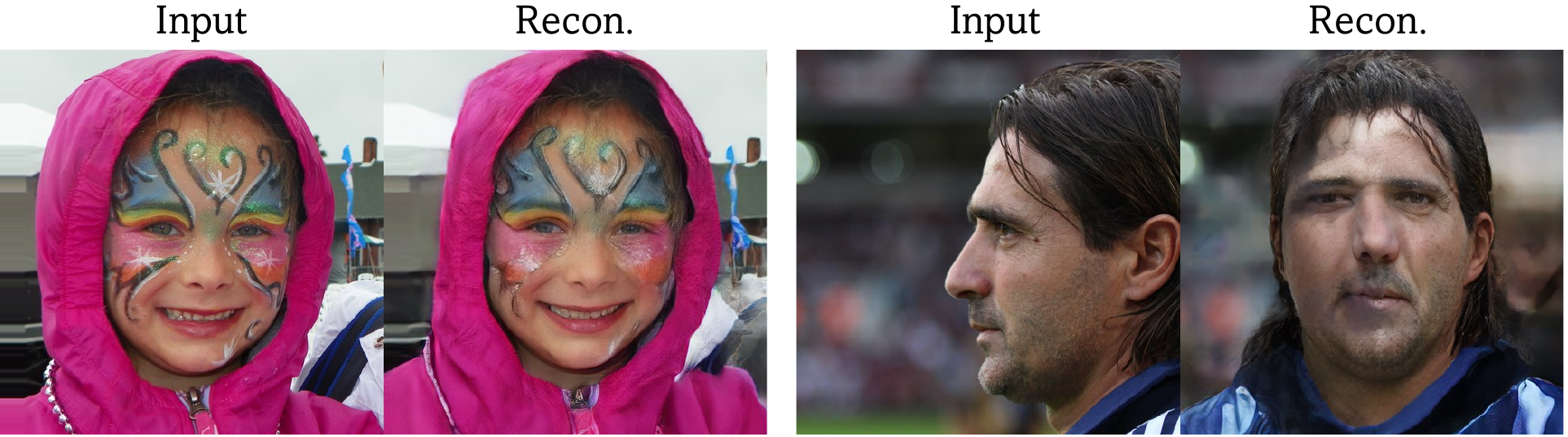}
	\caption{Limitations of our method. It can produce inferior results for input with heavy makeups and large viewing angles.}
	\label{fig:limitation}
 \vspace{-15pt}
\end{figure}

\subsection{Limitations and Future Works} While our method can synthesize high-fidelity 4D head avatar at a single shot, it still has some limitations.

\begin{figure*}[p]
    \small
	\centering
	\includegraphics[width=1.0\textwidth]{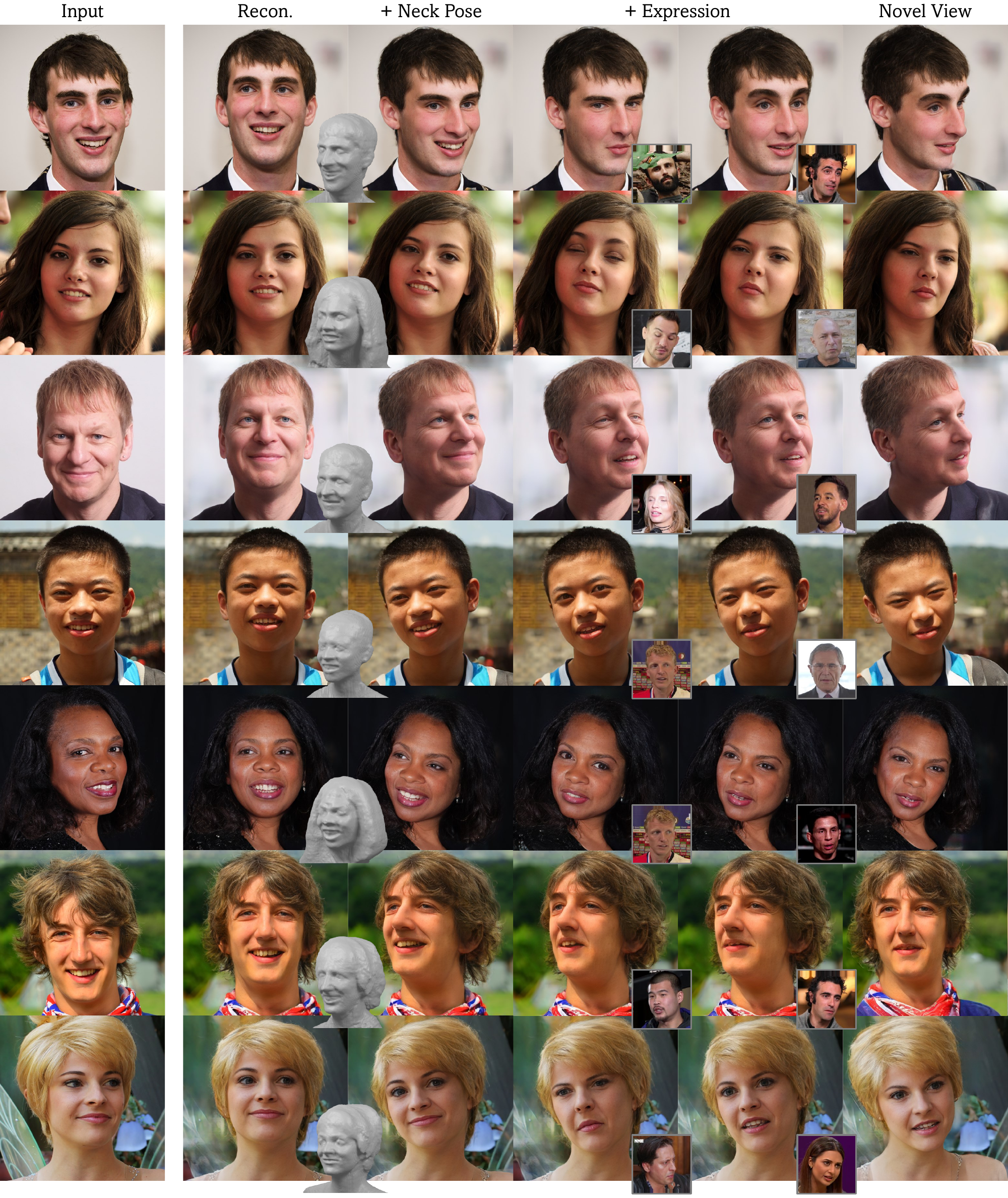}
	
	\caption{One-shot 4D head synthesis results by our method.\label{fig:results_main_supp1}}

% 	\vspace{-8pt}
\end{figure*}

Our method cannot well handle complex accessories and makeups as shown in Fig.~\ref{fig:limitation}. It also struggles to reconstruct high-frequency details in the background (\eg, last row in Fig.~\ref{fig:results_main_supp1}). We believe this problem can be mitigated by increasing the volume rendering resolution to allow for more intricate information flow. This would also help with the texture flickering issue brought by the 2D super-resolution module. Learning with synthetic data of more diverse appearance can also be helpful.

When the input image is nearly profile with large yaw angles, our method can produce inferior results due to out-of-distribution issue (see Fig.~\ref{fig:limitation}). We are also aware of artifacts under certain expressions such as eye blink, as the GenHead model for data synthesis is learned on FFHQ dataset with relatively less images of closed eyes. It is possible to leverage data with more diverse expressions and poses to improve the model's generalizability.

Currently, the synthetic data from GenHead relies on 3DMM for expression control which can be less vivid compared to that of real data, and thus restricts the motion control ability of our method. Besides, the data synthesis process requires training an animatable 3D-aware GAN in advance which is also challenging and can suffer from loss of modality issue of GAN. Learning on 4D synthetic data also encounters more severe overfitting issue compared to learning on static 3D data as in~\cite{trevithick2023real}. Therefore, synthesizing 4D data of better quality and diversity to facilitate the one-shot reconstruction pipeline is a key problem to be solved. Alternatively, it is worth exploring an effective way to incorporate real data and 3D priors into an end-to-end training framework. Apart from that, extending the current pipeline to support few-shot cases is also an important direction.

\subsection{Ethics Consideration} The goal of this paper is to create animatable head avatars for virtual communications. However, without a proper supervision, it can be misused for creating deceptive contents to people. We do not condone any such harmful behavior. Incorporation of advanced deepfake detectors~\cite{rossler2019faceforensics++,cozzolino2021id,corvi2023detection} is a possible way to prevent the potential misuse.

\end{document}